%% file: ms.tex
\documentclass{article}

\usepackage{PRIMEarxiv}

\usepackage[utf8]{inputenc} 
\usepackage[T1]{fontenc}    
\usepackage{hyperref}       
\usepackage{url}            
\usepackage{booktabs}       
\usepackage{amsfonts}       
\usepackage{amsmath}
\usepackage{amssymb}
\usepackage{tcolorbox}
\usepackage{nicefrac}       
\usepackage{microtype}      
\usepackage{lipsum}
\usepackage{fancyhdr}       
\usepackage{graphicx}       
\usepackage{xcolor}
\usepackage{soul}
\usepackage{listings}
\usepackage{mathptmx}
\usepackage{bm}
\usepackage{booktabs}
\usepackage{listings}
\usepackage{adjustbox}
\usepackage{subcaption}
\graphicspath{{media/}}     

\newcommand{\citep}[1]{\cite{{#1}}}
\newcommand{\upn}[0]{^{(n)}}

\newcommand{\R}[0]{\mathbb{R}}
\newcommand{\cut}[1]{}

\definecolor{codegreen}{rgb}{0,0.6,0}
\definecolor{codegray}{rgb}{0.5,0.5,0.5}
\definecolor{codepurple}{rgb}{0.58,0,0.82}
\definecolor{backcolour}{rgb}{1.0,1.0,1.0}

\lstdefinestyle{mystyle}{
    backgroundcolor=\color{backcolour},
    commentstyle=\color{codegreen},
    keywordstyle=\color{magenta},
    numberstyle=\tiny\color{codegray},
    stringstyle=\color{codepurple},
    basicstyle=\ttfamily\footnotesize,
    breakatwhitespace=false,
    breaklines=true,
    captionpos=b,
    keepspaces=true,
    numbers=left,
    numbersep=5pt,
    showspaces=false,
    showstringspaces=false,
    showtabs=false,
    tabsize=2,
    moredelim=[is][\underbar]{_}{_}
}

\newtheorem{definition}{Definition}[section]

\title{Torch-Choice: A PyTorch Package for Large-Scale Choice Modeling with Python
}

\author{
  Tianyu Du \\
  Stanford University \\
  \texttt{tianyudu@stanford.edu} \\
   \And
  Ayush Kanodia \\
  Stanford University \\
  \texttt{akanodia@stanford.edu}\thanks{Work primarily done during PhD at Stanford University} \\
  \And
  Susan Athey \\
  Stanford University \\
  \texttt{athey@stanford.edu} \\
}

\begin{document}
\maketitle

\begin{abstract}
\input{sections/abstract.tex}
\end{abstract}


\input{sections/introduction.tex}
\input{sections/defn_terms.tex}
\input{sections/data_management.tex}

\input{sections/models.tex}
\input{sections/benchmark.tex}
\input{sections/conclusion.tex}
\newpage \quad \newpage
\input{sections/acknowledgements.tex}
\bibliographystyle{unsrt}
\bibliography{refs}

\end{document}

%% file: sections/abstract.tex
The \code{torch-choice} is an open-source library for flexible, fast choice modeling with Python and PyTorch.
\code{torch-choice} provides a \code{ChoiceDataset} data structure to manage databases flexibly and memory-efficiently. The paper demonstrates constructing a \code{ChoiceDataset} from databases of various formats and functionalities of \code{ChoiceDataset}.
The package implements two widely used models, namely the multinomial logit and nested logit models, and supports regularization during model estimation.
\code{torch-choice} incorporates the option to take advantage of GPUs for estimation, allowing it to scale to massive datasets while being computationally efficient. Models can be initialized using either R-style formula strings or Python dictionaries.
We conclude with a comparison of the computational efficiencies of \code{torch-choice} and \code{mlogit} in R \cite{JSSv095i11} as (1) the number of observations increases, (2) the number of covariates increases, and (3) the expansion of item sets. Finally, we demonstrate the scalability of \code{torch-choice} on large-scale datasets. 

%% file: sections/introduction.tex
\section{Introduction}

Choice models are ubiquitous in economics, computer science, marketing, psychology, education and more fields.
In consumer demand modeling, the researcher considers a setting where a consumer selects a product or brand from a set of alternatives, and the goal is to analyze how product characteristics affect consumer purchase decisions. These estimates can in turn be used to answer counterfactual questions about how purchases would change if, for example, prices or availability change.  Choice models can be applied to aggregate market level demand data \citep{berry1995}, cross-sectional individual choice data \citep{berry2004}, and panel data where customer purchase events are tracked over time \citep{chintagunta2001}.
In transportation choice modeling, choice models are used to analyze intercity passenger travel mode choice \citep{mcfadden1974measurement, wilson1990}, and in spatial modeling to model location choice \citep{miyamoto2004}.
In labor economics, choice models are used to model labor supply and childcare decisions \citep{kornstad2007}, and family labor supply \citep{van1995}.
In education, Item Response Theory (IRT) and related models are choice models \citep{baker2001basics, embretson2013}.
In psychology, choice models are used to model decision-making under uncertainty  \citep{bonoma1979}, and psychiatric treatment diagnoses \citep{fleiss1986}.
In computer science, choice models are used to build recommendation systems \citep{zhang2002} as well as for classification tasks such as spam filtering \citep{chang2008}, image and character recognition \citep{deng2012mnist}, and many other applications.

Multinomial logistic regression and extensions are the most commonly used choice models. More recently, there has been work using choice models and their extensions (matrix-factorization, multilayer neural networks for choice modeling, Bayesian embeddings based choice models) to model restaurant choice \citep{athey2018}, grocery shopping \citep{donnelly2021}, inter-dependencies among choices when putting together a shopping basket \citep{ruiz2020}, the welfare effects of recommendations \citep{donnelly2022}, and to model labor market transitions \citep{vafa2022}.
  For each of the models, either models based on multinomial logistic regression are used, or they are important baselines for further model development, creating a strong need for effective choice modeling software.

Recent years have seen the proliferation of various tools and software packages that go a long way in making it easy to estimate the parameters of choice models. While each of these have their particular strengths, there are important limitations of these packages when they are used with the goal of building flexible, fast choice models which scale to massive data.

In the discussion that follows, when we refer to \code{R} packages, we mean \code{glm} and \code{mlogit}. When we refer to \code{Stata} packages, we mean the entire suite of choice models offered in the packaged software.\footnote{See \url{https://www.stata.com/manuals/cm.pdf}, \url{https://www.stata.com/help.cgi?xtmelogit} and \url{https://www.stata.com/manuals/cmnlogit.pdf}} When we refer to flexible functional forms, we mean model specifications with user- and item-specific latent coefficients (latent variables). When we refer to regularization, we mean the ability to incorporate penalization or ``shrinkage'' of regression coefficients \citep{krogh1991simple}.
\begin{enumerate}
    \item \code{R} packages allow the researcher to easily specify and learn models, and visualize the results, but they are not GPU-accelerated. Further, the estimation methods and data structures are not optimized for scalability (see Section \ref{section:performance}). \code{R} also offers limited forms of latent variables in the specification of utility for multinomial and nested logit; specifically under random coefficients specifications, it does not allow for the retrieval of user-specific latent coefficients. \code{R} does not support regularization of model parameters.
    \item \code{Stata} packages offer a similar user experience to \code{R}. They afford better computational efficiency by utilizing multiple CPU cores, although this is a paid feature, and can be prohibitively expensive for some users. However, these are still not GPU-accelerated features in \code{Stata}. Similar to \code{R}, while \code{Stata} supports user-specific latent coefficients in random coefficients models, the researcher must write additional code to access these coefficients, post estimation, while \code{torch-choice} provides these directly. \code{Stata} also does not support the regularization of model parameters.
    \item Writing one's own models in \code {PyTorch} has a steep learning curve for statisticians and econometricians, and setting up the training loop requires advanced programming expertise.
    \item In \code{python}, the \code{scikit-learn} Logistic Regression \citep{scikit-learn} package is the go-to tool for computer scientists and allows for regularization, but it does not allow for specifying flexible functional forms, nor does it allow the availability of items to change across sessions. The \code{pyBLP} package \citep{MicroPyBLP} implements BLP-type random coefficients logit models in Python, and \code{xlogit}\citep{arteaga2022xlogit} implements choice models with a focus on the multinomial logit specification; \code{torch-choice} offers more flexible utility functional forms, leverages PyTorch and allows for using GPUs for faster estimation.
\end{enumerate}

We propose \code{torch-choice}, a library for flexibly, fast choice modeling with PyTorch; it implements conditional and nested logit models, designed for both estimation and prediction. This package aims to address these limitations and fill the gap between tools used by econometricians/statisticians (R-style, end-to-end model training) and those computer scientists who are familiar with (PyTorch).

A limitation of \code{torch-choice}
is that while it natively supports estimating user-specific coefficients in the setting of \emph{panel data}, \code{torch-choice} estimates a constant coefficient (instead of distribution) for each user, and the same coefficient is applied across all sessions involving this user. It does not, however, learn the distribution over user-level parameters, and does not model that distribution.
As such, it does not support random coefficients in \emph{cross-sectional data} settings, where one can learn the parameters of distribution over user-level parameters in \code{R}, \code{Stata}, or \code{xlogit} in Python \citep{arteaga2022xlogit}. Please see section \ref{section:models} for more details.

What are the advantages of using \code{torch-choice}?
\begin{enumerate}
    \item It offers computational efficiency advantages because it is built on PyTorch, which allows for (optional) GPU acceleration.  With PyTorch, our packages easily scale to large datasets of millions of choice records. In contrast, existing packages do not leverage GPUs, and use only one CPU core by default. \code{Stata} allows for the use of multiple cores, but this is a paid feature.\footnote{\$ 375 per year for a 4 core student license, prices are higher for more cores. See \url{https://www.stata.com/order/new/edu/profplus/student-pricing/}.}
    \item It uses a custom data structure that improves efficiency in memory and storage. The \code{ChoiceDataset} data structure avoids storing duplicate covariate information and offers an advantage in memory usage compared to traditional long and wide formats of choice datasets. Moreover, Pytorch allows for easy switching between different numeric precision (e.g., from float32 to float16); such a feature future reduces the memory requirement on large-scale datasets.
    \item Model parameters are estimated using state-of-the-art first-order optimization algorithms such as Adam \citep{kingma2014adam}; this makes it run in linear time (in the number of parameters) as compared to the quadratic, Hessian-based optimization routines used for optimizing choice models in \code{R} and \code{Stata}. Further, one can use any optimizer from the collection of state-of-the-art optimizers by in PyTorch, which have been tested by numerous modern machine learning practitioners. \code{torch-choice} supports second-order Quasi-Newton methods such as Limited-memory BFGS (LBFGS) for more precise optimization when computational resources permit\citep{liu1989lbfgs}.
    \item \code{torch-choice}'s model estimation scheme is built upon PyTorch-Lightning\citep{falcon2019pytorch}, which facilitates rich model management such as learning rate decaying, early stopping, and parallel hyper-parameter tuning. These features are particularly useful when comparing multiple models over enormous datasets.
    Moreover, \code{torch-choice} complies TensorBoard logs during model estimation to illustrate the training progress, allowing for easy model diagnostics.
    \item It natively reports user-level parameters if the functional form is so specified; while this can be accomplished using choice modeling software in \code{R} and \code{Stata}, it requires additional programming and expertise.
    \item It allows the researcher to specify the functional form of utility (including specifying session-specific variables such as price), and to specify availability sets that may vary across sessions. This is not possible with \code{scikit-learn} and requires the researcher to write such functionality from scratch in standard \code{PyTorch}. Thus, \code{torch-choice} combines the benefits of scale obtained by using GPU acceleration with flexibility.
    \item It supports nested logit; this is also not possible with \code{scikit-learn} and vanilla \code{PyTorch}.\footnote{We support a two level nested logit model. We do not support multiple categories under nested logit at this time, this is left to future work.}
    \item It is open source, and can be easily customized and built upon. This is not so for choice models in \code{Stata}, though it is true for those in \code{R} and \code{scikit-learn}.
    \item When using the multinomial logit with multiple categories, it allows sharing of parameters across categories \citep{rudolph2017structured}, which is not supported in \code{R}, \code{Stata}, or \code{Python}.
\end{enumerate}

The \code{torch-choice} library is publicly available on Github at: \href{https://github.com/gsbDBI/torch-choice}{\code{torch-choice} github repository}.
Researchers can install our packages via pip following instructions at \href{https://pypi.org/project/torch-choice}{\code{torch-choice} on pip}.
We have prepared tutorials on the documentation website: \href{https://gsbdbi.github.io/torch-choice}{\code{torch-choice} documentation website}, where researchers can find the full documentation of APIs.
Jupyter notebook \citep{Kluyver2016jupyter} versions of these tutorials are available on Github at: \href{https://github.com/gsbDBI/torch-choice/tree/main/tutorials}{\code{torch-choice tutorial notebooks}}. Code demonstrated in this paper can be found in this \href{https://github.com/gsbDBI/torch-choice/blob/main/tutorials/paper_demo.ipynb}{Jupyter Notebook}.

Section 2 outlines the choice problem \code{torch-choice} was built to model.
Section 3 covers the data structures. Section 4 covers the models implemented in \code{torch-choice}. Section 5 shows performance benchmarks for \code{torch-choice} against another open source package, \code{mlogit} in \code{R}, and demonstrates its performance advantages. Finally, Section 6 concludes this paper.

%% file: sections/defn_terms.tex
\section{The Choice Problem}
We start by outlining the building blocks of the choice problem that will be processed with \code{torch-choice}.
Models in \code{torch-choice} predict \emph{which item among all available items a user will choose given the context (where the context is referred to as a \emph{session})}.
The paper refers to the basic unit of analysis as a choice record.

\begin{figure}[!ht]
    \centering
    \includegraphics[width=1.0\textwidth]{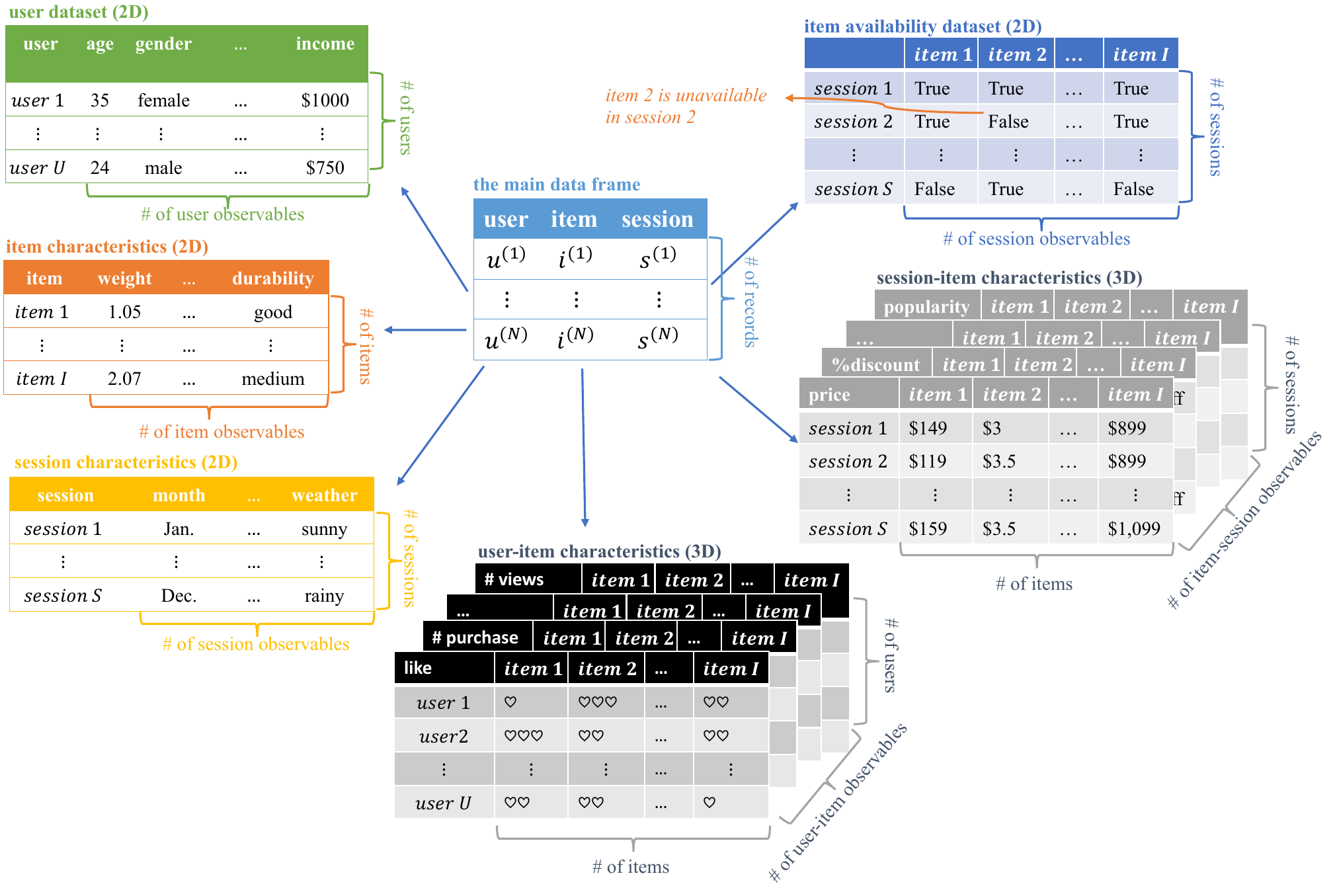}
    \caption{Illustration of \code{ChoiceDataset} Data Structure. The \code{torch-choice} supports user-session-item-specific observables as well, which is a 4D tensor. We are not drawing it in this illustration for succinctness.}
    \label{fig:choice-dataset-illustration}
\end{figure}

\begin{definition}[Choice Records]
    A \textbf{choice record}, also referred to as simply a \textbf{record}, is a triplet (\textbf{user}, \textbf{item-chosen}, \textbf{session}), where \textbf{user} is the identity of the user who made a choice, \textbf{item chosen} is the item the user chose, and an index for the \textbf{session}, where a session is characterized by contextual information such as the time-varying characteristics of the choice alternatives and the identities of available choice alternatives.
\end{definition}
Figure~\ref{fig:choice-dataset-illustration} illustrates the layout of the \code{ChoiceDataset} data structure.
Each row of the central dataframe in Figure \ref{fig:choice-dataset-illustration} corresponds to a choice record.

\begin{table}[h]
    \centering
    \resizebox{\textwidth}{!}{
        \begin{tabular}{c|c}
            \toprule
            Symbol & Definition \\
            \midrule
            $N$ & Number of choice records \\
            $n$ & Index of choice record, $n \in \{1, 2, \dots, N\}$ \\
            $U, I, S, C$ & Number of users/items/sessions/categories \\
            $u\upn, i\upn, s\upn$ & User, item chosen, and session index of the $n^{th}$ choice record \\
            $\mathcal{C}(i)$ & Category of item $i$, $\mathcal{C}(i) \in \{1, 2, \dots, C\}$ \\
            $I_{c}$ & Items in category $c$, $I_c \subseteq \{1, 2, \dots, I\}$ \\
            $I_{\mathcal{C}(i)}$ & Items in the same category of item $i$ \\
            $\mathcal{D}$ & The entire dataset $\{(u\upn, i\upn, s\upn)\}_{n=1}^N$ \\
            $\mathcal{D}\upn$ & The $n^{th}$ record in dataset $\mathcal{D}$, $(u\upn, i\upn, s\upn)$ \\
            $A$ & The $S \times I$ binary matrices indicating the availability of items in each session \\
            $A(s)$ & The subset of items available in session $s$ \\
            $\mathcal{V}_{uis}$ & The utility user $u$ gains from item $i$ in session $s$ \\
            $\mu_{uis}$ & The deterministic part of utility $\mathcal{V}_{uis} = \mu_{uis} + \varepsilon_{uis}$ \\
            $\varepsilon_{uis}$ & The stochastic part of utility $\mathcal{V}_{uis} = \mu_{uis} + \varepsilon_{uis}$ \\
            $P(i \mid u, s)$ & The probability that user $u$ chooses item $i$ among items in $\mathcal{C}(i)$ in session $s$ \\
            $\hat{P}(i \mid u, s)$ & The predicted probability that user $u$ chooses item $i$ among items in $\mathcal{C}(i)$ in session $s$ \\
            \bottomrule
        \end{tabular}
    }
    \caption{Summary of Notations}
    \label{tab:notation-summary}
\end{table}

\subsection{Users, Items, and Sessions}
Choice records are the fundamental unit of observation in choice modeling describing \emph{\textbf{who} \textit{(i.e., which user)} chooses \textbf{what} \textit{(i.e., which item)} under what context \textit{(i.e., in which session)}}.
Let $N$ denote the size of the dataset and $n \in \{1, 2, \dots, N\}$ denote the index for records in the dataset.
Let $\mathcal{D}$ denote the dataset and  $\square\upn$, the super-script with parentheses, denote the $n$-th record in the dataset, where $\square$ is a placeholder for the object of interest.
Therefore, $\mathcal{D}\upn = (u\upn, i\upn, s\upn)$ denotes the item chosen, the user involved, and the corresponding session index of the $n$-th record.
Similarly, to represent subsets of a dataset, we can use superscripts. For instance, $\mathcal{D}^{(\text{train})}$ denotes the training set and $\mathcal{D}^{(\text{test})}$ denotes the test set.

The set of $U$ \textbf{users} is indexed by $u \in \{1, 2, \dots, U\}$, while the set of $I$ \textbf{items} is indexed by $i \in \{1, 2, \dots, I\}$. The set of \textbf{sessions} $S$ is indexed by $s \in \{1, 2, \dots, S\}$.
The set of items is partitioned into $C$ \textbf{categories} indexed by $c \in \{1,2,\dots,C\}$.\footnote{The data structure in \code{torch-choice} does not limit how many categories a user could choose each time. The notion of categories is fully general and can be any partition of the item set. However, the nested-logit model implemented in \code{torch-choice} allows for modeling only one category, with two levels of nesting. See the model section for more details. However, modeling multiple categories in a single model is not currently supported for nested-logit models implemented in \code{torch-choice}. This is left to future work.}
Let $I_c$ denote the collection of items in category $c$, $I_c$'s are disjoint sets and $\bigcup_{c=1}^C I_c = \{1, 2, \dots I\}$. Note that when we learn over multiple categories simultaneously, parameters that are not item-specific
are the same across categories \citep{rudolph2017structured}. To learn per category parameters at the user and constant level, one needs to specify separate datasets and learn separate models over them.
Let $\mathcal{C}(i) \in \{1, 2, \dots, C\}$ denote the category of item $i$, $I_{\mathcal{C}(i)}$ represents items in the same category as item $i$.\footnote{Since we will be using PyTorch to train our model, we represent user, item, session identities with \emph{consecutive integer values} instead of the raw human-readable names of items (e.g., Dell 24-inch LCD monitor). Similarly, we encode user indices and session indices as well.
Raw item names can be encoded to integers easily with \href{https://scikit-learn.org/stable/modules/generated/sklearn.preprocessing.LabelEncoder.html}{\code{sklearn.preprocessing.LabelEncoder}} or \href{https://scikit-learn.org/stable/modules/generated/sklearn.preprocessing.OrdinalEncoder.html}{\code{sklearn.preprocessing.OrdinalEncoder}}.}

The model assumes each user must choose exactly one item from each category (where one item may be defined to represent not selecting any items) and that choices are independent across categories and conditional on user preferences.
The user $u$ chooses an item from each category to \emph{maximize} their utility $\mathcal{V}_{uis}$ from item $i$ in the context of session $s$. We assume the utility can be decomposed into a deterministic utility $\mu_{uis}$ and an error term $\varepsilon_{uis}$ following the Gumbel distribution as in Equation~(\ref{eq:utility-decomposition}).\footnote{The Gumbel distribution is a continuous distribution with a single parameter $\beta$ and the probability density function is given by $f(x) = \frac{1}{\beta} \exp(-x/\beta - \exp(-x/\beta))$. It is easy to show that with Gumbel noise, the probability density of $\max_{i}\mathcal{V}_{uis}$ follows the softmax function of $\mu_{uis}$.}

\begin{equation}
    \mathcal{V}_{uis} = \mu_{uis} + \varepsilon_{uis}
    \label{eq:utility-decomposition}
\end{equation}


The same user will make choices \emph{from different categories} in the same session, so that the dataset will contain multiple purchasing records with the same indices for user and session $(u, s)$ but different item indices $i$.

The utility $\mu_{uis}$ depends on the characteristics of users and items and other factors (e.g., the utility from iced coffee drops during winter time). We introduce the notion of \textbf{session} indexed by $s \in \{1,2,\dots, S\}$ to capture factors that vary by time and context.

\begin{tcolorbox}
\textbf{Example} Suppose a researcher collected data from a single store for over a year. In this case, we can define the session $s$ as the date of purchase and $S = 365$.
The notion of sessions can be more flexible; for example, if we aggregated data from multiple stores and want to distinguish between online orders and in-store purchases, we can define the session as \code{(date, storeID, IsOnlineOrder)} instead.
\end{tcolorbox}
The flexibility of sessions enables easier data management (e.g., observables and item availability) as we will see later.

\subsection{Item Availability and Flexible Choice Sets}
The session can be used to capture the fact that item availability may change over time.
Our dataset admits a $S \times I$ binary matrix $A$ to indicate which items are available to be selected in each session.\footnote{The item availability acts as a constraint in users' choice problems. Computationally, we assign $-\infty$ utility to unavailable items to operationalize their exclusion in the choice expression.}
We use $A(s)$ to denote the set of available items in session $s$. All items are available all the time by default, i.e., $A(s) = \{1, 2, \dots, I\}$ for all session $s$.
Researchers should define the granularity of this matrix guided by the granularity with which item availability changes, along with the granularity of how users make decisions. For example, if the availability of items is the same in a day and a user faces at most one choice decision in a day, sessions may be defined at a day level; however, if users make more than one choice per day or if the availability of items is personalized to users, sessions will need to be more granular. Less granular session definitions encourage more efficient data management computationally. Note that in some applications, the prominence with which an item is displayed can also change over time; that can be captured as a time-varying item characteristic (i.e., an item-session observable).

\subsection{Observable Tensors}
The objective of our models is to predict utilities $\mu_{uis}$ as a function of user, item, and session characteristics called \textbf{observable tensors}.\footnote{Observables are also referred to as \textbf{features} in machine learning literature, \textbf{independent variables} in statistics, they are often denoted as $X$ with appropriate super/sub-script.}

\begin{equation}
    \mu_{uis} = f_\text{model}(W_u, X_i, Z_{ui}, T_s, P_{is}, M_{uis}).
\end{equation}

The package supports all six types of observables with various dependencies on user $u$, item $i$, and session $s$. Observables are essentially high-dimensional arrays called tensors in machine learning literature. For example, an array $X$ of shape $U \times I \times 3$ can encapsulate three user-item-specific variables, indexing \code{X[u,i,:]} retrieves a vector in $\R^3$ for these observables corresponding to user \code{u} and item \code{i}.
The following is a list of observable types supported and expected shapes of tensors/arrays encapsulating them.

\begin{enumerate}
    \item \textbf{User} observable tensors $\in \mathbb{R}^{U\times K_\text{user}}$ vary \emph{only} with the user identity, for example, time-invariant user demographics.

    \item \textbf{Item} observable tensors $\in \mathbb{R}^{I\times K_\text{item}}$ vary \emph{only} with the item index, for example, the length of a movie.

    \item \textbf{Session} observable tensors $\in \mathbb{R}^{S \times K_\text{session}}$ vary \emph{only} with the session identifier. For example, if the session is defined as the date, the daily weather could be a session observable.

    \item \textbf{User-item} observable tensors $\in \mathbb{R}^{U \times I \times K_\text{user-item}}$ depend on both user and item. For example, a user-item session observable can capture users' ratings of different items, assuming users cannot change their ratings. User-item observable tensors can also capture interaction effects between users and items.

    \item \textbf{Item-session} observable tensors $\in \mathbb{R}^{S \times I \times K_\text{item-session}}$ are session-and-item-specific observables that vary with \emph{both} session and item, for example, prices of items if sessions are defined as dates.

    \item \textbf{User-session-item} observable tensors $\in \mathbb{R}^{U \times S \times I \times K_\text{user-session-item}}$ are specific to the combination of user, item, and session. For example, a user-session-item observable can store the number of times a user purchased an item prior to a specific date.
\end{enumerate}


\textbf{Note on Computational Efficiency} The user-session-item observable tensor can be extremely large and cause out-of-memory issues even at moderate levels of $U$, $I$, and $S$. For example, suppose we are modeling shopping data with customers as users and dates as sessions; there is a user-session-item-specific observable $x_{uis}$. Storing the complete $x_{uis}$ requires $U \times S \times I$ entries. Often, each customer only shops on a couple of dates, say three dates on average and $S \gg 3$. We can instead define sessions to be all existing pairs of (customer, date), and $X$ becomes a session-item-specific observable instead. Storing $X$ as a session-item-observable only requires $3 \times S \times I$ entries, a factor of $U/3$ reduction.


Since \code{torch-choice} leverages PyTorch and GPU accelerations, all observables need to be numerical, and indices of users/items/sessions should be \emph{consecutive} integers starting from 0.
Raw item names can be encoded easily with \href{https://scikit-learn.org/stable/modules/generated/sklearn.preprocessing.LabelEncoder.html}{\code{sklearn.preprocessing.LabelEncoder}} or \href{https://scikit-learn.org/stable/modules/generated/sklearn.preprocessing.OrdinalEncoder.html}{\code{OrdinalEncoder}}.

\begin{table}[!h]
    \centering
    \begin{tabular}{c|c|c}
    \toprule
        user\_index & session\_index & item\_index \\ \midrule
        Amy & Sep-17-2021-Store-A & banana \\
        Ben & Sep-17-2021-Store-B & apple \\
        Ben & Sep-16-2021-Store-A & orange \\
        Charlie & Sep-16-2021-Store-B & apple \\
        Charlie & Sep-16-2021-Store-B & orange \\
    \bottomrule
    \end{tabular}
    \caption{The Example Dataset}
    \label{tab:integer-encoding-example}
\end{table}
\begin{tcolorbox}
    \textbf{Example} Suppose we have a dataset of purchase history described in Table \ref{tab:integer-encoding-example} from two stores (Store A and B) on two dates (Sep 16 and 17), both stores sell \code{\{apple, banana, orange\}} and there are three people who came to those stores between Sep 16 and 17.

    The user, item, and session indices are strings in this example, which should be encoded to integers in pre-processing.
    After encoding, the above dataset becomes \code{user\_index=[0,1,1,2,2]}, \code{session\_index=[0,1,2,3,3]}, and \code{item\_index=[0,1,2,1,2]}. Users are encoded into \code{0=Amy, 1=Ben, 2=Charlie}, items are encoded into \code{0=banana, 1=apple, 2=orange}, and sessions are encoded into \code{0=Sep-17-2021-Store-A, 1=Sep-17-2021-Store-B, 2=Sep-16-2021-Store-A, 3=Sep-16-2021-Store-B}.
    Suppose we believe people's purchasing decision depends on the nutrition levels of these fruits; suppose apple has the highest nutrition level and banana has the lowest one, we can add
    \code{item\_obs=[[1.5], [12.0], [3.3]]} $\in \mathbb{R}^{3\times 1}$ with shape number-of-items by number-of-observable.
\end{tcolorbox}


\subsection{Model Prediction}
\input{sections/short_model_intro.tex}

The models implemented in \code{torch-choice} offer flexible ways to estimate the deterministic part $\mu_{uis}$.
Given the Gumbel distribution of $\varepsilon_{uis}$ and the independence of irrelevant items assumption, the predicted probability for choosing item $i$ among all items in the same category $I_{\mathcal{C}(i)}$ is a function of utilities $\{\mu_{ui's}\}_{i' \in I_{\mathcal{C}(i)}}$ from the model.
The exact functional form differs by model, and we defer the discussion to the model section.
The form of predicted probabilities for any given specificaton of $\mu$ is a normalized exponential function (``softmax'') function in Equation~(\ref{eq:prob-choice}).
\begin{equation}
    P(i | u, s) = \begin{cases}
        \frac{\exp\left(\mu_{uis}\right)}{\sum_{i' \in I_{\mathcal{C}(i)} \cap A(s)} \exp\left(\mu_{ui's}\right)} & \text{if $i \in A(s)$, i.e., $i$ is available in session $s$.}\\
        0 &\text{otherwise}
    \end{cases}
    \label{eq:prob-choice}
\end{equation}




%% file: sections/short_model_intro.tex
We assume each user selects at most one item from each item category each session by maximizing his/her utility $\mathcal{V}_{uis}$ within each category.
When users are choosing over items, we can write utility $\mathcal{V}_{uis}$ that user $u$ derives from item $i$ in session $s$ as:
\begin{equation}
    \mathcal{V}_{uis} = \mu_{uis} + \varepsilon_{uis}
\end{equation}

where $\varepsilon$ is an unobserved random error term.
If we assume i.i.d. extreme value type I errors for $\varepsilon_{uis}$, which leads to the logistic probabilities of user $u$ choosing item $i$ among $I_{C(i)}$ (i.e., items in the same category as $i$)  in session $s$, as shown by McFadden\citep{mcfadden1973conditional}, and as often studied in Econometrics.

Given the distributional assumption on $\varepsilon_{uis}$, the probability that user $u$ chooses item $i$ in session $s$ is captured by the logistic function
\begin{equation}
    P(i\mid us) = \frac{\exp\left(\mu_{uis}\right)}{\sum_{\ell \in A(s) \cap I_{C(i)}}\exp\left(\mu_{u\ell s}\right)}
    \label{eq:logistic-choice-probability}
\end{equation}
where $A(s)$ is the set of items available in session $s$ (i.e., the choice set), and $I_{C(i)}$ is the set of items in the same category as item $i$. The probability in Equation \eqref{eq:logistic-choice-probability} is normalized over available items only; therefore, probabilities of choosing unavailable item are zero. The mean utility $\mu_{uis}$ for a user, item, and session is specified according to a functional form, and we consider different combinations of observable and latent variables, each of which corresponds to a distinct version of the model.

Equation \eqref{eq:clm-general-form} describes the general form of the utility function in conditional logit models. Greek letters with tilde and prime (e.g., $\alpha, \tilde{\alpha}_i, \alpha_u'$) are coefficients to be estimated; capital letters denote observables; subscripts of them denote their dependencies on user $u$, item $i$, and session $s$.

\begin{align}
\begin{split}
    \mu_{uis} &= \underbrace{\alpha + \tilde{\alpha}_i + \alpha_u'}_\text{fixed effects}
    + \underbrace{\beta W_u^{\text{I}} + \tilde{\beta}_i W_u^{\text{II}} + \beta_u' W_u^{\text{III}}}_{\text{user observables}}
    + \underbrace{\gamma X_i^{\text{I}} + \tilde{\gamma}_i X_i^{\text{II}} + \gamma_u' X_i^{\text{III}}}_{\text{item observables}}
    + \underbrace{\eta Z_{ui}^\text{I} + \tilde{\eta}_i Z_{ui}^\text{II} + \eta_u' Z_{ui}^\text{III}}_\text{user-item observables}
    \\
    &+ \underbrace{\tau T_s^{\text{I}} + \tilde{\tau}_i T_s^{\text{II}} + \tau_u' T_s^{\text{III}}}_{\text{session observables}}
    + \underbrace{\delta P_{is}^{\text{I}} + \tilde{\delta}_i P_{is}^{\text{II}} + \delta_u' P_{is}^{\text{III}}}_{\text{item-session observables}}
    + \underbrace{\omega M_{uis}^{\text{I}} + \tilde{\omega}_i M_{uis}^{\text{II}} + \omega'_i M_{uis}^{\text{III}}}_\text{user-session-item observables}
\end{split}
\label{eq:clm-general-form}
\end{align}

For example, the specification in Equation \eqref{eq:clm-simple-example} allows for user observables ($W_u$), item observables ($X_i$), and item-session observable ($P_{is}$), and intercepts that vary with the item:
\begin{equation}
    \mu_{uis} = \alpha_i + \beta_u X_i + \gamma W_u + \delta_i P_{is}
    \label{eq:clm-simple-example}
\end{equation}
The Greek letters $\alpha_i$, $\beta_u$, $\gamma$, and $\delta_i$ denote coefficients to be estimated. These coefficients can be interpreted as the individual user fixed effect, impacts of item characteristics, user characteristics, and (session, item)-specific characteristics on the mean utility for user $u$ for item $i$.
Researchers can add user-item observable interactions by constructing the corresponding item-session specific observable.
The \code{torch-choice} package offers flexible ways to estimate the mean utility $\mu_{uis}$ beyond this example; the model section will cover these aspects in detail.

Beyond the basic and widely implemented multinomial logit model, the \code{torch-choice} library provides two widely used classes of choice models, the \emph{conditional logit model} (CLM) and the \emph{nested logit model} (NLM) to model $\mu_{uis}$. The standard multinomial logit model is nested in the conditional logit model, and so is treated as a special case. The nested logit model relaxes the i.i.d. assumption on $\varepsilon_{uis}$.

In each of the conditional logit and nested logit models, researchers can specify the functional form they want to use for observables.
Each model takes user and session information $(u\upn, s\upn)$ together with observables and outputs the predicted probability of purchasing each $\ell \in \{1, 2, \dots, I\}$, denoted as $\hat{P}(\ell \mid u\upn, s\upn)$. In cases when not all items are available, the model sets the probability of unavailable items to zero and normalizes probabilities among available items.
By default, model coefficients are estimated using the gradient descent algorithm to maximize log-likelihood in Equation~(\ref{eq:general-log-likelihood}).
Since \code{torch-choice} is built upon PyTorch, any other optimization algorithms supported by PyTorch can be used as well. Please refer to our online documentation for details regarding applying alternative optimization algorithms.

\begin{equation}
    \text{log-likelihood} = \sum_{n=1}^N \log \hat{P}(i\upn \mid u\upn, s\upn)
    \label{eq:general-log-likelihood}
\end{equation}

%% file: sections/data_management.tex
\section{Data Structures}

This section covers the \code{ChoiceDataset} class in the \code{torch-choice} library.\footnote{Please see our online documentation \href{https://gsbdbi.github.io/torch-choice/data_management/}{Data Management Tutorial} and \href{https://gsbdbi.github.io/torch-choice/easy_data_management/}{Easy Data Wrapper Tutorial} for more details. We provide Jupyter notebooks offering hands on experience on our \href{https://github.com/gsbDBI/torch-choice/tree/main/tutorials}{Github repository}.}\footnote{Code examples in this section are for illustrating the usage only; please refer to online materials for executable notebooks.}
\code{ChoiceDataset} is a specialized instantiation of PyTorch's \code{Dataset} class. It inherits support for all of the functionality of PyTorch's \code{Dataset} class, including subsetting, indexing, and random sampling.
Most importantly, GPU memory is often much more scarce on modern machines, \code{ChoiceDataset} is manage information in a memory-efficient manner. Existing packages such as \code{xlogit} in Python \cite{arteaga2022xlogit} and \code{mlogit} \cite{JSSv095i11} in R require the data to be in a long-format or wide-format, which store duplicate information.
\code{ChoiceDataset} minimizes its memory footage via its specialized structure eliminating redundant information.
Moreover, all information in \code{ChoiceDataset} is stored as PyTorch tensors,\footnote{Tensors are simply PyTorch's implementation of multi-dimensional arrays. Tensor operations are very similar to operations on \code{numpy} arrays; the \href{https://pytorch.org/tutorials/beginner/blitz/tensor_tutorial.html}{PyTorch tensor tutorial} overviews basics of tensors. This manuscript doesn't assume researchers' knowledge on PyTorch tensor operations. Two-dimensional tensors are simply matrices; we will use tensor and matrix interchangeably in this manuscript.} which allows researchers to leverage half-precision floats to further improve memory efficiency and seamless data transfer between CPU and GPU.

\subsection{Constructing a Choice Dataset, Method 1: Easy Data Wrapper Class}


The \code{torch-choice} package provides two methods to build \code{ChoiceDataset}; the first method is to use the \code{EasyDataWrapper} class, which is a wrapper class that converts pandas \code{DataFrame} in long-format to \code{ChoiceDataset}.

\input{sections/easy_data.tex}

\subsection{Constructing a Choice Dataset, Method 2: Building from Tensors}


As discussed above, storing data in a long format is not memory efficient due to repeated data. \code{torch-choice} enables storage of observables as separate tensors.

This section demonstrates an example of creating a dataset from observable and index tensors. We use a randomly generated dataset with $N=10,000$ records from $U=10$ users, $I=4$ items and $S=500$ sessions.
We generate random observable tensors for user, item, session, and item-session observables from independent multivariate Gaussian distributions following Equation~(\ref{eq:random-obs-data}).\footnote{We chose 128, 64, 10, and 12 as the number of observables, this choice is arbitrary and researchers can choose any number of observables.}
For simplicity, we generate data where all items are available in all sessions.
\begin{align}
    \begin{aligned}
    W_u &\sim \mathcal{N}(\bm{0}, I_{128 \times 128}) \quad \forall u \in \{1, 2, \dots, U\} \\
    X_i &\sim \mathcal{N}(\bm{0}, I_{64 \times 64}) \quad \forall i \in \{1, 2, \dots, I\} \\
    Z_{ui} &\sim \mathcal{N}(\bm{0}, I_{32 \times 32}) \quad \forall (u, i) \in \{1, 2, \dots, U\} \times \{1, 2, \dots, I\} \\
    T_s &\sim \mathcal{N}(\bm{0}, I_{10 \times 10}) \quad \forall s \in \{1, 2, \dots, S\} \\
    P_{is} &\sim \mathcal{N}(\bm{0}, I_{12 \times 12}) \quad \forall (i, s) \in \{1, 2, \dots, I\} \times \{1, 2, \dots, S\} \\
    B_{us} &\sim \mathcal{N}(\bm{0}, I_{10 \times 10}) \quad \forall (u, s) \in \{1, 2, \dots, U\} \times \{1, 2, \dots, S\} \\
    M_{uis} &\sim \mathcal{N}(\bm{0}, I_{8 \times 8}) \quad \forall (u, i, s) \in \{1, 2, \dots, U\} \times \{1, 2, \dots, I\} \times \{1, 2, \dots, S\} \\
    u\upn &\sim \text{Unif}(\{1, 2, \dots, U\}) \quad \forall n \in \{1, 2, \dots, N\} \\
    i\upn &\sim \text{Unif}(\{1, 2, \dots, I\}) \quad \forall n \in \{1, 2, \dots, N\} \\
    s\upn &\sim \text{Unif}(\{1, 2, \dots, S\}) \quad \forall n \in \{1, 2, \dots, N\}
    \end{aligned}
    \label{eq:random-obs-data}
\end{align}

The code snippet below builds a \code{ChoiceDataset} object from randomly generated  $(u\upn, i\upn, s\upn)$ and observable tensors. This construction assumes that all items are available in all sessions, since the \code{item_availability} argument is not specified.
\begin{lstlisting}[language=python]
N = 10_000
num_users = 10
num_items = 4
num_sessions = 500

user_obs = torch.randn(num_users, 128)
item_obs = torch.randn(num_items, 64)
useritem_obs = torch.randn(num_users, num_items, 32)
session_obs = torch.randn(num_sessions, 10)
itemsession_obs = torch.randn(num_sessions, num_items, 12)
usersession_obs = torch.randn(num_users, num_sessions, 10)
usersessionitem_obs = torch.randn(num_users, num_sessions, num_items, 8)

item_index = torch.LongTensor(np.random.choice(num_items, size=N))
user_index = torch.LongTensor(np.random.choice(num_users, size=N))
session_index = torch.LongTensor(np.random.choice(num_sessions, size=N))
item_availability = torch.ones(num_sessions, num_items).bool()

dataset = ChoiceDataset(
    # pre-specified keywords of __init__
    item_index=item_index,  # required.
    num_items=num_items,
    # optional:
    user_index=user_index,
    num_users=num_users,
    session_index=session_index,
    item_availability=item_availability,
    # additional keywords of __init__
    user_obs=user_obs,
    item_obs=item_obs,
    session_obs=session_obs,
    itemsession_obs=itemsession_obs,
    useritem_obs=useritem_obs,
    usersession_obs=usersession_obs,
    usersessionitem_obs=usersessionitem_obs)
\end{lstlisting}

The \code{ChoiceDataset} is expecting the following arguments at its initialization.
\begin{enumerate}
    \item \code{item\_index} captures item chosen in each record $(i\upn)_{n=1}^N$.
    \item \code{user\_index} captures the user making decision in each record $(u\upn)_{n=1}^N$. \code{user\_index} is optional. By default, the same user is making decisions repeatedly.
    \item \code{session\_index} captures the corresponding session of each record $(s\upn)_{n=1}^N$. \code{session\_index} is optional. By default, each record has its own session.
    \item \code{item\_availability} is the binary availability matrix $A$ with shape $S \times I$. \code{item\_availability} is optional. By default, all items are available in all sessions.
\end{enumerate}


\begin{table}[h]
    \centering
    \resizebox{\textwidth}{!}{
        \begin{tabular}{c|c|c}
            \toprule
            observable type & tensor shape & keyword argument prefix requirement \\
            \midrule
            item-specific & ($I$, number of observables) & \code{item\_<obs\_name>} \\
            user-specific & ($U$, number of observables) & \code{user\_<obs\_name>} \\
            (user, item)-specific & ($U$, $I$, number of observables) & \code{useritem\_<obs\_name>} \\
            session-specific & ($S$, number of observables) & \code{session\_<obs\_name>}\\
            (item, session)-specific & ($S$, $I$, number of observables) & \code{itemsession\_<obs\_name>} \\
             (user, session)-specific & ($U$, $S$, number of observables) & \code{usersession\_<obs\_name>} \\
            (user, session, item)-specific & ($U$, $S$, $I$, number of observables) & \code{usersessionitem\_<obs\_name>} \\
            \bottomrule
        \end{tabular}
    }
    \medbreak
    \caption{Expected Shapes of Observable Tensors. Dimensions of tensors are always ordered as user dimension first, then session dimension, and then item dimension.}
    \label{tab:tensor-shape}
\end{table}

As mentioned previously, all observable tensors should meet shape requirements in Table \ref{tab:tensor-shape}.

Further, the data structure in \code{torch-choice} package utilizes variable name prefixes to keep track of observables' variation. Observable tensors need to follow naming conventions for \code{ChoiceDataset} to build correctly.
While constructing the \code{ChoiceDataset} with observable tensor keyword arguments, observable tensors are named with prefixes indicating how they depend on the user (with \code{user\_} prefix), item (with \code{item\_} prefix), session (with \code{session\_} prefix), and both of (item, session) (with \code{itemsession\_} prefix) .

For example, suppose we have a tensor \code{X} for user income of shape $(U, 1)$. The income observable should be introduced as \code{ChoiceDataset(..., user\_income=X, ...)}, initializing the dataset with \code{ChoiceDataset(..., income\_of\_user=X, ...)} using an incorrect keyword argument will introduce errors.
Table \ref{tab:tensor-shape} summarizes the naming requirement for different types of observable tensors.

When multiple observable tensors of the same level of variation are needed, researchers can include as many keyword arguments as needed, as long as they follow the prefix requirement. For example, \code{ChoiceDataset(..., user\_income=X, user\_age=Z, user\_education=E...)} includes three different user-specific observable tensors. Such a feature is particularly useful when you want, for example, an item-specific effect on users' income but a constant effect on users' education levels.


\subsection{Functionalities of the Choice Dataset}
This section overviews the functionalities of the \code{ChoiceDataset} class. The command \code{print(dataset)} provides a quick overview of shapes of tensors included in the object as well as where the dataset is located (i.e., in CPU memory or GPU memory). The \code{dataset.summary()} method offers more detailed information about the dataset.

\begin{lstlisting}[language=python]
print(f"{dataset=:}")
# dataset=ChoiceDataset(num_items=4, num_users=10, num_sessions=500, label=[], item_index=[10000], user_index=[10000], session_index=[10000], item_availability=[500, 4], user_obs=[10, 128], item_obs=[4, 64], session_obs=[500, 10], itemsession_obs=[500, 4, 12], useritem_obs=[10, 4, 32], usersession_obs=[10, 500, 10], usersessionitem_obs=[10, 500, 4, 8], device=cpu)
\end{lstlisting}

The \code{dataset.num\_\{users, items, sessions\}} attribute can be used to obtain the number of users, items, and sessions, which are determined automatically from the \code{\{user, item, session\}\_obs} tensors provided while initializing the dataset object. The \code{len(dataset)} reveals the number of records $N$ in the dataset.

\begin{lstlisting}[language=python]
print(f'{dataset.num_users=:}')
# dataset.num_users=10
print(f'{dataset.num_items=:}')
# dataset.num_items=4
print(f'{dataset.num_sessions=:}')
# dataset.num_sessions=500
print(f'{len(dataset)=:}')
# len(dataset)=10000
\end{lstlisting}

The \code{ChoiceDataset} offers a \code{clone} method to make a copy of the dataset; researchers can modify cloned dataset safely without changing the original dataset.
\begin{lstlisting}[language=python]
# clone
print(f"{dataset.item_index[:10]=:}")
# dataset.item_index[:10]=tensor([2, 3, 0, 2, 2, 3, 0, 0, 2, 1])
dataset_cloned = dataset.clone()
# modify the cloned dataset.
dataset_cloned.item_index = 99 * torch.ones(num_sessions)
print(f"{dataset_cloned.item_index[:10]=:}")
# dataset_cloned.item_index[:10]=tensor([99., 99., 99., 99., 99., 99., 99., 99., 99., 99.])
# the cloned dataset is changed.
print(f"{dataset.item_index[:10]=:}")
# dataset.item_index[:10]=tensor([2, 3, 0, 2, 2, 3, 0, 0, 2, 1])
\end{lstlisting}

\code{ChoiceDataset} objects can be moved between host memory (i.e., CPU memory) and device memory (i.e., GPU memory) using the \code{dataset.to()} method.
The \code{dataset.to()} method moves all tensors in the \code{ChoiceDataset} to the target device.
The following code runs only on a machine with a compatible GPU as well as a GPU-compatible version of PyTorch installed.
The \code{dataset.\_check\_device\_consistency()} method checks if all tensors are on the same device.
For example, an attempt to move the \code{item_index} tensor to CPU without moving other tensors will result in an error message.

\begin{lstlisting}[language=python]
# move to device
print(f'{dataset.device=:}')
# dataset.device=cpu
print(f'{dataset.device=:}')
# dataset.device=cpu
print(f'{dataset.user_index.device=:}')
# dataset.user_index.device=cpu
print(f'{dataset.session_index.device=:}')
# dataset.session_index.device=cpu

dataset = dataset.to('cuda')

print(f'{dataset.device=:}')
# dataset.device=cuda:0
print(f'{dataset.item_index.device=:}')
# dataset.item_index.device=cuda:0
print(f'{dataset.user_index.device=:}')
# dataset.user_index.device=cuda:0
print(f'{dataset.session_index.device=:}')
# dataset.session_index.device=cuda:0
\end{lstlisting}

\begin{lstlisting}[language=python]
dataset._check_device_consistency()
\end{lstlisting}

\begin{lstlisting}[language=python]
# NOTE: this cell will result errors, this is intentional.
dataset.item_index = dataset.item_index.to('cpu')
dataset._check_device_consistency()
# ---------------------------------------------------------------------------

# Exception                                 Traceback (most recent call last)

# <ipython-input-56-40d626c6d436> in <module>
        # 1 # # NOTE: this cell will result errors, this is intentional.
        # 2 dataset.item_index = dataset.item_index.to('cpu')
# ----> 3 dataset._check_device_consistency()

# ~/Development/torch-choice/torch_choice/data/choice_dataset.py in _check_device_consistency(self)
    # 180                 devices.append(val.device)
    # 181         if len(set(devices)) > 1:
# --> 182             raise Exception(f'Found tensors on different devices: {set(devices)}.',
    # 183                             'Use dataset.to() method to align devices.')
    # 184

# Exception: ("Found tensors on different devices: {device(type='cuda', index=0), device(type='cpu')}.", 'Use dataset.to() method to align devices.')
\end{lstlisting}

The \code{ChoiceDataset.x\_dict} attribute reconstructs the long-format representation of the dataset, which is useful for data exploration, visualization, and designing customized models upon \code{ChoiceDataset}.
\code{dataset.x\_dict} is a dictionary mapping observable names to tensors with shape \code{(N, num\_items, *)}, where \code{N} is the length of dataset, \code{num\_items} is the number of items, and \code{*} is the dimension of the corresponding observable.
The helper function \code{print\_dict\_shape} prints the shapes of tensors in a dictionary.

\begin{lstlisting}[language=python]
# collapse to a dictionary object.
print_dict_shape(dataset.x_dict)
# dict.user_obs.shape=torch.Size([10000, 4, 128])
# dict.item_obs.shape=torch.Size([10000, 4, 64])
# dict.session_obs.shape=torch.Size([10000, 4, 10])
# dict.itemsession_obs.shape=torch.Size([10000, 4, 12])
# dict.useritem_obs.shape=torch.Size([10000, 4, 32])
# dict.usersession_obs.shape=torch.Size([10000, 4, 10])
# dict.usersessionitem_obs.shape=torch.Size([10000, 4, 8])
\end{lstlisting}

 \code{dataset[indices]} can be applied with \code{indices} as an integer-valued tensor or array to retrieve the corresponding rows of the dataset.\footnote{In Python, the square bracket indexing operation implicitly calls the \code{\_\_getitem\_\_} method of the object. For example, \code{dataset[indices]} is equivalent to \code{dataset.\_\_getitem\_\_(indices)}, we occasionally call the subset method as \code{\_\_getitem\_\_}.}
 The subset \code{ChoiceDataset} contains \code{item\_index}, \code{user\_index}, and \code{session\_index} corresponding to $(u\upn, i\upn, s\upn)_{n \in \text{selected indices}}$ and full copies of observable tensors.
The sub-setting method allows researchers to easily split dataset into training, validation, and testing subsets.

The example code block below randomly queries five records from the dataset.
\begin{lstlisting}[language=python]
# __getitem__ to get batch.
# pick 5 random sessions as the mini-batch.
dataset = dataset.to('cpu')
indices = torch.Tensor(np.random.choice(len(dataset), size=5, replace=False)).long()
print(f"{indices=:}")
# indices=tensor([7119, 9650, 5466, 1073, 8419])
subset = dataset[indices]
print(f"{dataset=:}")
# dataset=ChoiceDataset(num_items=4, num_users=10, num_sessions=500, label=[], item_index=[10000], user_index=[10000], session_index=[10000], item_availability=[500, 4], user_obs=[10, 128], item_obs=[4, 64], session_obs=[500, 10], itemsession_obs=[500, 4, 12], useritem_obs=[10, 4, 32], usersession_obs=[10, 500, 10], usersessionitem_obs=[10, 500, 4, 8], device=cpu)
print(f"{subset=:}")
# subset=ChoiceDataset(num_items=4, num_users=10, num_sessions=500, label=[], item_index=[5], user_index=[5], session_index=[5], item_availability=[500, 4], user_obs=[10, 128], item_obs=[4, 64], session_obs=[500, 10], itemsession_obs=[500, 4, 12], useritem_obs=[10, 4, 32], usersession_obs=[10, 500, 10], usersessionitem_obs=[10, 500, 4, 8], device=cpu)
\end{lstlisting}

It is worth noting that the subset method automatically creates a clone of the datasets so that any modification applied to the subset will not be reflected on the original dataset.
The code block below demonstrates this feature.

\begin{lstlisting}[language=python]
print("Before modifying the batch:")
print(f"{subset.item_index=:}")
print(f"{dataset.item_index[indices]=:}")

subset.item_index += 1  # modifying the batch does not change the original dataset.
print("After modifying the batch:")
print(f"{subset.item_index=:}")
print(f"{dataset.item_index[indices]=:}")

# Before modifying the batch:
# subset.item_index=tensor([2, 2, 1, 2, 3])
# dataset.item_index[indices]=tensor([2, 2, 1, 2, 3])
# After modifying the batch:
# subset.item_index=tensor([3, 3, 2, 3, 4])
# dataset.item_index[indices]=tensor([2, 2, 1, 2, 3])
\end{lstlisting}

\begin{lstlisting}[language=python]
print("Before modifying the batch:")
print(f"{subset.item_obs[0, 0]=:}")
print(f"{dataset.item_obs[0, 0]=:}")

print("After modifying the batch:")
subset.item_obs += 1
print(f"{subset.item_obs[0, 0]=:}")
print(f"{dataset.item_obs[0, 0]=:}")

# Before modifying the batch:
# subset.item_obs[0, 0]=-0.294853538274765
# dataset.item_obs[0, 0]=-0.294853538274765
# After modifying the batch:
# subset.item_obs[0, 0]=0.7051464319229126
# dataset.item_obs[0, 0]=-0.294853538274765
\end{lstlisting}

\begin{lstlisting}[language=python]
# these two are different objects in memory, indicated by the different memory addresses queried by id().
# note that the memory address of the same object will be different when the code is run again.
print(f"{id(subset.item_index)=:}")
print(f"{id(dataset.item_index[indices])=:}")
# id(subset.item_index)=128489592094128
# id(dataset.item_index[indices])=128489590644624
\end{lstlisting}

\subsection{Chaining Multiple Datasets with JointDataset}
The nested logit model requires two separate datasets, one for the categorical level information and the other for the item level information.
The \code{JointDataset} class offers a simple way to combine multiple \code{ChoiceDataset} objects. The following code snippet demonstrates how to create a joint dataset from two separate datasets, in which one dataset is named as \code{item\_level\_dataset} and the other one is named as \code{cateogry\_level\_dataset}. Researchers can name these datasets arbitrarily as long as names are distinct.

\begin{lstlisting}[language=python]
item_level_dataset = dataset.clone()
nest_level_dataset = dataset.clone()
joint_dataset = JointDataset(
    item=item_level_dataset,
    nest=nest_level_dataset)

print(f"{joint_dataset=:}")
"""
joint_dataset=JointDataset with 2 sub-datasets: (
	item: ChoiceDataset(num_items=4, num_users=10, num_sessions=500, label=[], item_index=[10000], user_index=[10000], session_index=[10000], item_availability=[500, 4], user_obs=[10, 128], item_obs=[4, 64], session_obs=[500, 10], itemsession_obs=[500, 4, 12], useritem_obs=[10, 4, 32], usersession_obs=[10, 500, 10], usersessionitem_obs=[10, 500, 4, 8], device=cpu)
	nest: ChoiceDataset(num_items=4, num_users=10, num_sessions=500, label=[], item_index=[10000], user_index=[10000], session_index=[10000], item_availability=[500, 4], user_obs=[10, 128], item_obs=[4, 64], session_obs=[500, 10], itemsession_obs=[500, 4, 12], useritem_obs=[10, 4, 32], usersession_obs=[10, 500, 10], usersessionitem_obs=[10, 500, 4, 8], device=cpu)
)
"""
\end{lstlisting}
The joint dataset structure allows easy and consistent indexing on multiple datasets.
The indexing method of \code{JointDataset} will return a dictionary of subsets of each dataset encompassed in the joint dataset structure.
Specifically, \code{joint\_dataset[indices]} returns a dictionary \code{\{"item\_level\_dataset": item\_level\_dataset[indices], "category\_level\_dataset": category\_level\_dataset[indices]\}}.
The chaining functionality is particularly helpful when the researcher wishes to experiment with the model on multiple datasets.

\subsection{Using Pytorch data loader for the training loop}
The \code{ChoiceDataset} is a subclass of PyTorch's dataset class and researchers with PyTorch expertise can use the PyTorch data loader to iterate over the dataset with customized batch size and shuffling options using PyTorch's \code{Sampler} and \code{DataLoader}.\footnote{This section covers advanced materials for researchers who want to build models upon \code{ChoiceDataset}. The \code{torch-choice} offers utilities for model estimation and researchers can safely skip this section.}
This section is intended for researchers who wish to develop their model training pipeline; researchers who are interested in the standard training loop provided by \code{torch-choice} and \code{bemb} can skip this section.

For demonstration purposes, we turned off the shuffling option. We first create an index sampler and then use it to create a data loader.

\begin{lstlisting}[language=python]
from torch.utils.data.sampler import BatchSampler, SequentialSampler, RandomSampler
shuffle = False  # for demonstration purpose.
batch_size = 32

# Create sampler.
sampler = BatchSampler(
    RandomSampler(dataset) if shuffle else SequentialSampler(dataset),
    batch_size=batch_size,
    drop_last=False)

dataloader = torch.utils.data.DataLoader(dataset,
                                         sampler=sampler,
                                         collate_fn=lambda x: x[0],
                                         pin_memory=(dataset.device == 'cpu'))
\end{lstlisting}

\begin{lstlisting}[language=python]
print(f'{item_obs.shape=:}')
# item_obs.shape=torch.Size([4, 64])
item_obs_all = item_obs.view(1, num_items, -1).expand(len(dataset), -1, -1)
item_obs_all = item_obs_all.to(dataset.device)
item_index_all = item_index.to(dataset.device)
print(f'{item_obs_all.shape=:}')
# item_obs_all.shape=torch.Size([10000, 4, 64])
\end{lstlisting}

The following code snippet iterates through the \code{dataloader} to get the mini-batches.
\begin{lstlisting}[language=python]
for i, batch in enumerate(dataloader):
    first, last = i * batch_size, min(len(dataset), (i + 1) * batch_size)
    idx = torch.arange(first, last)
    assert torch.all(item_obs_all[idx, :, :] == batch.x_dict['item_obs'])
    assert torch.all(item_index_all[idx] == batch.item_index)
\end{lstlisting}

\begin{lstlisting}[language=python]
batch.x_dict['item_obs'].shape
# torch.Size([16, 4, 64])
\end{lstlisting}

\begin{lstlisting}[language=python]
print_dict_shape(dataset.x_dict)
# dict.user_obs.shape=torch.Size([10000, 4, 128])
# dict.item_obs.shape=torch.Size([10000, 4, 64])
# dict.session_obs.shape=torch.Size([10000, 4, 10])
# dict.itemsession_obs.shape=torch.Size([10000, 4, 12])
# dict.useritem_obs.shape=torch.Size([10000, 4, 32])
# dict.usersession_obs.shape=torch.Size([10000, 4, 10])
# dict.usersessionitem_obs.shape=torch.Size([10000, 4, 8])
\end{lstlisting}

\begin{lstlisting}[language=python]
dataset.__len__()
# 10000
\end{lstlisting}

%% file: sections/easy_data.tex
The \code{EasyDataWrapper} only requires that the researcher (1) load dataframes to Python (for example, using \code{pandas} tools to load various types of data files including CSV, TSV, Stata database, and Excel spreadsheet) and (2) usage of \code{pandas} to pre-process the dataset.

We load the car-choice dataset to demonstrate \code{torch-choice}'s data wrapper utility. The car-choice dataset is a synthetic dataset on consumers' choices regarding the nationality of cars; the dataset was modified from Stata's tutorial dataset.
\begin{lstlisting}[language=python]
car_choice = pd.read_csv("https://raw.githubusercontent.com/gsbDBI/torch-choice/main/tutorials/public_datasets/car_choice.csv")
\end{lstlisting}

The dataset is in ``long'' format: multiple rows constitute a single choice record.

Table~\ref{tab:carchoice-df-head} presents the first 30 rows of the car-choice dataset, in which each user (consumer) only made one choice.
The first four rows with \code{record\_id == 1} correspond to the first \emph{purchasing record}.
It means that consumer 1 was deciding on four types of cars (i.e., \emph{items}) and chose \emph{American} car (since the \code{purchase == 1} in that row of \emph{American} car).
Not all cars were available all the time. For example, there is no row in the dataset with \code{session\_id == 4 \& car == "Korean"}, which means only American, Japanese, and European cars were available in the fourth session.

\begin{table}[ht]
    \centering
    \caption{The first 30 rows of the car-choice dataset.}
    \label{tab:carchoice-df-head}
    \medbreak
    \small
    \resizebox*{\textwidth}{!}{
    \begin{tabular}{rrrlrrrrrr}
    \toprule
     record\_id &  session\_id &  consumer\_id &      car &  purchase &  gender &    income &  speed &  discount &  price \\
    \midrule
             1 &           1 &            1 & American &         1 &       1 & 46.699997 &     10 &      0.94 &     90 \\
             1 &           1 &            1 & Japanese &         0 &       1 & 46.699997 &      8 &      0.94 &    110 \\
             1 &           1 &            1 & European &         0 &       1 & 46.699997 &      7 &      0.94 &     50 \\
             1 &           1 &            1 &   Korean &         0 &       1 & 46.699997 &      8 &      0.94 &     10 \\
             2 &           2 &            2 & American &         1 &       1 & 26.100000 &     10 &      0.95 &    100 \\
             2 &           2 &            2 & Japanese &         0 &       1 & 26.100000 &      8 &      0.95 &     70 \\
             2 &           2 &            2 & European &         0 &       1 & 26.100000 &      7 &      0.95 &     20 \\
             2 &           2 &            2 &   Korean &         0 &       1 & 26.100000 &      8 &      0.95 &     10 \\
             3 &           3 &            3 & American &         0 &       1 & 32.700000 &     10 &      0.90 &     80 \\
             3 &           3 &            3 & Japanese &         1 &       1 & 32.700000 &      8 &      0.90 &     60 \\
             3 &           3 &            3 & European &         0 &       1 & 32.700000 &      7 &      0.90 &     20 \\
             4 &           4 &            4 & American &         1 &       0 & 49.199997 &     10 &      0.81 &     50 \\
             4 &           4 &            4 & Japanese &         0 &       0 & 49.199997 &      8 &      0.81 &     40 \\
             4 &           4 &            4 & European &         0 &       0 & 49.199997 &      7 &      0.81 &     30 \\
             5 &           5 &            5 & American &         0 &       1 & 24.300000 &     10 &      0.87 &     80 \\
             5 &           5 &            5 & Japanese &         0 &       1 & 24.300000 &      8 &      0.87 &     30 \\
             5 &           5 &            5 & European &         1 &       1 & 24.300000 &      7 &      0.87 &     30 \\
             6 &           6 &            6 & American &         1 &       0 & 39.000000 &     10 &      0.83 &    100 \\
             6 &           6 &            6 & Japanese &         0 &       0 & 39.000000 &      8 &      0.83 &     60 \\
             6 &           6 &            6 & European &         0 &       0 & 39.000000 &      7 &      0.83 &     10 \\
             7 &           7 &            7 & American &         0 &       1 & 33.000000 &     10 &      0.98 &    100 \\
             7 &           7 &            7 & Japanese &         0 &       1 & 33.000000 &      8 &      0.98 &     60 \\
             7 &           7 &            7 & European &         1 &       1 & 33.000000 &      7 &      0.98 &     40 \\
             7 &           7 &            7 &   Korean &         0 &       1 & 33.000000 &      8 &      0.98 &     10 \\
             8 &           8 &            8 & American &         1 &       1 & 20.300000 &     10 &      0.88 &     60 \\
             8 &           8 &            8 & Japanese &         0 &       1 & 20.300000 &      8 &      0.88 &     50 \\
             8 &           8 &            8 & European &         0 &       1 & 20.300000 &      7 &      0.88 &     30 \\
             9 &           9 &            9 & American &         0 &       1 & 38.000000 &     10 &      0.93 &     90 \\
             9 &           9 &            9 & Japanese &         1 &       1 & 38.000000 &      8 &      0.93 &     90 \\
             9 &           9 &            9 & European &         0 &       1 & 38.000000 &      7 &      0.93 &     20 \\
    \bottomrule
    \end{tabular}
    }
\end{table}



The \code{EasyDataWrapper} requires a long-format \textbf{main dataset} with the following columns:
\begin{enumerate}
    \item
    \code{purchase\_record\_column}: a column identifies \textbf{record} (also called \textbf{case} in Stata syntax). In this tutorial, the \code{record\_id} column is the identifier. For example, the first 4 rows of the dataset (see above) has \code{record\_id == 1}, which implies that the first 4 rows together constitute the first record.
    \item
    \code{item\_name\_column}: a column identifies \textbf{names} of items, which is \code{car} in the dataset above. This column provides information above the availability as well. As mentioned above, there is no column with \code{car == "Korean"} in the fourth session (\code{session\_id == 4}), indicating that the Korean car was not available in that session.
    \item
    \code{choice\_column}: a column with values $\in \{0, 1\}$ identifies the \textbf{choice} made by the consumer in each record, which is the \code{purchase} column in the car choice dataset. There should be exactly one row per record (i.e., rows with the same values in \code{purchase\_record\_column}) with value one, while the values are zeros for all other rows.
    \item
    \code{user\_index\_column}: an \emph{optional} column identifies the \textbf{user} making the choice, which is \code{consumer\_id} in the car-choice dataset.
    \item
    \code{session\_index\_column}: an \emph{optional} column identifies the \textbf{session} of choice, which is \code{session\_id} in the above dataframe.
\end{enumerate}



\subsubsection{Adding Observables}
The next step is to identify observables to be incorporated in the model.
In the car-choice dataset, we want to add
 (1) \code{gender} and \code{income} columns as user-specific observables; (2) \code{speed} as item-specific observable; (3) \code{discount} as session-specific observable; (4) \code{price} as (session, item)-specific observable.

\subsubsection{Adding Observables, Method 1: Observables Derived from Columns of the Main Dataset}
The above car-choice dataframe already includes observables we want to add.
The first approach to adding observables is to simply identify these columns while initializing the \code{EasyDatasetWrapper} object. This is accomplished by supplying a list of column names to each of \code{\{user, item, session, itemsession\}\_observable\_columns} keyword arguments.
For example, we assign
\begin{lstlisting}[language=python]
user_observable_columns=["gender", "income"]
\end{lstlisting}
to specify user-specific observables from the \code{gender} and \code{income} columns of the main dataframe.
The code snippet below shows how to initialize the \code{EasyDatasetWrapper} object.


\begin{lstlisting}[language=python]
from torch_choice.utils.easy_data_wrapper import EasyDatasetWrapper
data_wrapper_from_columns = EasyDatasetWrapper(
    main_data=car_choice,
    purchase_record_column='record_id',
    choice_column='purchase',
    item_name_column='car',
    user_index_column='consumer_id',
    session_index_column='session_id',
    user_observable_columns=['gender', 'income'],
    item_observable_columns=['speed'],
    session_observable_columns=['discount'],
    itemsession_observable_columns=['price'])

data_wrapper_from_columns.summary()
dataset = data_wrapper_from_columns.choice_dataset
print(f"{dataset=:}")
# dataset=ChoiceDataset(num_items=4, num_users=885, num_sessions=885, label=[], item_index=[885], user_index=[885], session_index=[885], item_availability=[885, 4], item_speed=[4, 1], user_gender=[885, 1], user_income=[885, 1], session_discount=[885, 1], itemsession_price=[885, 4, 1], device=cpu)
\end{lstlisting}

The \code{data_wrapper.summary()} method can be used to print out a summary of the dataset.
The \code{ChoiceDataset} object can be accessed by the wrapper using \code{data_wrapper.choice_dataset}.

\subsubsection{Adding Observables, Method 2: Added as Separated DataFrames}
Researchers can also construct dataframes and use dataframes to supply different observables. This is useful in the case of longitudinal data or when users choose items from multiple categories, so that there are multiple choice records by the same user. Suppose there are $K$ choice records for \emph{each} of the $U$ users.
Using a single dataframe for all variables requires a lot of memory and repeated data, while using a separate dataframe mapping user index to user observables only requires $U$ entries for each observable.
Similarly, storing item information or item-session information in long-format data leads to inefficient storage.

Dataframes for user-specific observables need to contain a column of user IDs (e.g., \code{consumer\_id}), this column should have exactly the same name as the column containing user indices.
Similarly, for item-specific observables, the dataframe should contain a column of item names (i.e., \code{car} in this car-choice example); session identifier column \code{session\_id} needs to be present for session-specific observable dataframes, and both item and session identifiers need to be present for item-session-specific observable dataframes.

The following code snippet extracts dataframes for observables from the car-choice dataframe.
\begin{lstlisting}[language=python]
# create dataframes for gender and income. The dataframe for user-specific observable needs to have the `consumer_id` column.
gender = car_choice.groupby('consumer_id')['gender'].first().reset_index()
income = car_choice.groupby('consumer_id')['income'].first().reset_index()
# alternatively, put gender and income in the same dataframe.
gender_and_income = car_choice.groupby('consumer_id')[['gender', 'income']].first().reset_index()
# speed as item observable, the dataframe requires a `car` column.
speed = car_choice.groupby('car')['speed'].first().reset_index()
# discount as session observable. the dataframe requires a `session_id` column.
discount = car_choice.groupby('session_id')['discount'].first().reset_index()
# create the price as itemsession observable, the dataframe requires both `car` and `session_id` columns.
price = car_choice[['car', 'session_id', 'price']]
# fill in NANs for (session, item) pairs that the item was not available in that session.
price = price.pivot(index='car', columns='session_id', values='price').melt(ignore_index=False).reset_index()
\end{lstlisting}

Then, researchers can initialize the \code{EasyDatasetWrapper} object with these dataframes via \code{\{user, item, session, itemsession\}\_observable\_data} keyword arguments.

\begin{lstlisting}[language=python]
data_wrapper_from_dataframes = EasyDatasetWrapper(
    main_data=car_choice,
    purchase_record_column='record_id',
    choice_column='purchase',
    item_name_column='car',
    user_index_column='consumer_id',
    session_index_column='session_id',
    user_observable_data={'gender': gender, 'income': income},
    # alternatively, supply gender and income as a single dataframe.
    # user_observable_data={'gender_and_income': gender_and_income},
    item_observable_data={'speed': speed},
    session_observable_data={'discount': discount},
    itemsession_observable_data={'price': price})

# the second method creates exactly the same ChoiceDataset as the previous method.
assert data_wrapper_from_dataframes.choice_dataset == data_wrapper_from_columns.choice_dataset
\end{lstlisting}

The \code{EasyDataWrapper} also support the mixture of above methods: in the following example, we supply user, item, and session-specific observables (i.e., gender, income, speed, and discount) as dataframes but the item-session specific observable (i.e., price) using the corresponding column name.

\begin{lstlisting}[language=python]
data_wrapper_mixed = EasyDatasetWrapper(
    main_data=car_choice,
    purchase_record_column='record_id',
    choice_column='purchase',
    item_name_column='car',
    user_index_column='consumer_id',
    session_index_column='session_id',
    user_observable_data={'gender': gender, 'income': income},
    item_observable_data={'speed': speed},
    session_observable_data={'discount': discount},
    itemsession_observable_columns=['price'])

# these methods create exactly the same choice dataset.
assert data_wrapper_mixed.choice_dataset == data_wrapper_from_columns.choice_dataset == data_wrapper_from_dataframes.choice_dataset
\end{lstlisting}

%% file: sections/models.tex
\section{Models in Torch-Choice Package}\label{section:models}
The \code{torch-choice} package implements two classes of models, namely the Conditional Logit Models (CLM) and Nested Logit Models (NLM).
Researchers specify the utility as a function of coefficients and observables.
There are multiple ways to handle user random coefficients in existing packages; for example in \code{Stata}, one can model a distribution $\mathcal{D}$ and treat user-specific coefficients as samples from this distribution, $\theta_u \sim \mathcal{D}$, and \code{Stata} then learns the parameters of the distribution $\mathcal{D}$.
The \code{torch-choice} package estimates a real-valued coefficient for every user, $\bm{\theta}_u$, instead of a distribution of coefficients over all users. This is useful when learning user level coefficients in panel data settings, and an advantage of \code{torch-choice} is that in these settings, it directly reports per user coefficients.\footnote{In contrast, users need to write their own code to estimate these per user coefficients in \code{Stata}.} In contrast, since \code{torch-choice} does not support learning a distribution over user level parameters, it cannot be used to learn a random effects model with cross-sectional data, since user level parameters would not be well identified in this case with only one record per user.
For example, suppose the researcher specifies a functional form $\mu_{uis} = \theta_u \times \text{Price}_i$, models in \code{torch-choice} will estimate $U$ scalar coefficients, one $\theta_u \in \R$ for each user $u$.
The coefficient specific to user $u$ is estimated using all choice records from user $u$, and the coefficient is the same across all sessions in which user $u$ is involved.
The researcher can use the model's summary method to obtain coefficients estimated for all users.
\input{sections/conditional_logit_model.tex}
\input{sections/nested_logit_model.tex}

%% file: sections/conditional_logit_model.tex
\subsection{Conditional Logit Model}
The CLM admits a linear form of the utility function.
Equation~(\ref{eq:genearl-utility-clm}) demonstrates the general form of the utility function for the conditional logit model, in which each $\bm{\theta}_p$ is a latent coefficient of the model to be estimated, and $\bm{x}_p$ is an observable vector.
The coefficient $\bm{\theta}_p$ can be fixed for all $(u, i, s)$, user-specific, or item-specific and the observable $\bm{x}_p$ can be user-specific, item-specific, session-specific, or (session, item)-specific.
Please note that the researcher can set $\bm{x}_1 = 1 \in \R$ to include an intercept term (including user-fixed effect and item-fixed effect) in the utility function.

\begin{align}
    \mu_{uis}
    \label{eq:genearl-utility-clm} = \sum_{p=1}^P \bm{\theta}_p^\top \bm{x}_p
\end{align}

The CLM assumes that all items are in the same category, and the probability for user $u$ to choose item $i$ under session $s$ is given by the logistic function in Equation~(\ref{eq:clm-softmax}).

\begin{equation}
    P(i \mid u, s) = \frac{\exp(\mu_{uis})}{\sum_{\ell \in A(s)} \exp(\mu_{u\ell s})}
    \label{eq:clm-softmax}
\end{equation}

One notable property of the CLM is the independence of irrelevant alternatives (IIA). The IIA property states that the relative probability for choosing item $i$ and $j$ is independent from any other item, which is evident from Equation~(\ref{eq:clm-iia}).

\begin{equation}
    \frac{P(i \mid u, s)}{P(j \mid u, s)} = \frac{\exp(\mu_{uis})}{\exp(\mu_{ujs})}
    \label{eq:clm-iia}
\end{equation}

This section demonstrates the usage of CLM model with the \emph{Mode-Canada} transportation choice dataset \cite{bhat1995heteroscedastic-modecanada}
The transportation choice dataset contains 3,880 travelers' choices of traveling method for the Montreal-Toronto corridor.
The dataset includes the cost, frequency, in-vehicle and out-vehicle time (IVT and OVT) of four transportation methods (air, bus, car, and train), and individual income as observables.

We assign a separate session to each choice record for the transportation dataset.
The cost, frequency, OVT, and IVT are session-item-specific observables.
The income of travelers only depends on the session.\footnote{We can also include the income variable as an user-specific observable. However, having income as a session-specific observable is more flexible and allows for different income levels of the same user but in different sessions.}
We model the utility as the following.
\begin{equation}
    \mu_{uis} = \lambda_i + \beta^{\top} X^{itemsession: (cost, freq, ovt)}_{is} + \gamma_i X^{session:income}_s + \delta_i X_{is}^{itemsession:ivt} + \varepsilon_{uis}
    \label{eq:clm-example-formula}
\end{equation}
where $X^{itemsession: (cost, freq, ovt)}_{is} \in \R^3$ denotes the vector of cost, frequency, and in-vehicle time of transportation choice $i$ in session $s$. $X^{session:income}_s \in \R$ denotes the income of decision maker in session $s$, and $X_{is}^{itemsession:ivt} \in \R$ represents the in-vehicle time of choice $i$ in session $s$.

\subsubsection{Initialize CLM with R-like Formula and Dataset}
The \code{torch-choice} offers two methods for creating the model.
Researchers can specify the model in Equation \eqref{eq:clm-example-formula} with a \code{formula} similar to that used in the R programming language, as follows:
\begin{lstlisting}[language=python]
# load Mode Canada transportation dataset
from torch_choice.data import load_mode_canada_dataset
dataset = load_mode_canada_dataset()
model = ConditionalLogitModel(
    formula='(itemsession_cost_freq_ovt|constant) + (session_income|item) + (itemsession_ivt|item-full) + (intercept|item)',
    dataset=dataset,
    num_items=4)
\end{lstlisting}
More generally, the \code{formula} string specifies the functional form of $\mu_{uis}$ as additive terms \code{(observable|variation)} as the following.
\begin{lstlisting}[language=python]
formula = "(observable_1|variation_1) + (observable_2|variation_2) + (observable_3|variation_3i) + ..."
\end{lstlisting}
The \code{observable\_i} is the observable name (i.e., name of $\bm{x}_p$ in \code{ChoiceDataset}),
The \code{variation\_i} specifies how the coefficient (i.e., $\bm{\theta}_p$) depends on the user, item, and session.
The \code{variation} takes the following possible values.
The coefficient can be (1) \code{constant} for all users and items; (2) \code{user} fits a different coefficient for each user; (3) \code{item-full} setting fits a different coefficient for each item (i.e., the item with index 0); or (4) \code{item} configuration is similar to the \code{item-full} setting, but the coefficient to the first item will be set to zero.
The researcher needs to pass in the \code{dataset} so that \code{model} can infer dimensions of $\bm{x}_p$'s.

\subsubsection{Initialize CLM with Dictionaries}
Alternatively, the code snippet below shows how to create the \code{ConditionalLogitModel} class with Python dictionaries.
The dictionary-based method is more suitable if researchers want to create multiple model configurations systematically.

\begin{lstlisting}[language=python]
model = ConditionalLogitModel(
    coef_variation_dict={'itemsession_cost_freq_ovt': 'constant',
                         'session_income': 'item',
                         'itemsession_ivt': 'item-full',
                         'intercept': 'item'},
    num_param_dict={'itemsession_cost_freq_ovt': 3,
                    'session_income': 1,
                    'itemsession_ivt': 1,
                    'intercept': 1},
    num_items=4)
\end{lstlisting}

The researcher needs to specify three keyword arguments while initializing the \code{ConditionalLogitModel} class.
To fit a CLM on \code{ChoiceDataset}, there should be an attribute of the choice dataset for every \emph{observable} $\bm{x}_p$.
For each observable, the user can specify the coefficient variation (i.e., the coefficient type) and the number of parameters for the observable, which is the same as the dimension of the observable due to the inner product.

\begin{enumerate}
    \item \code{coef\_variation\_dict} maps each observable name to the variation of coefficient associated to the observable.
    \item \code{num\_param\_dict} specifies the number of parameters for each coefficient, equivalently, the dimension of the observable.
    \item \code{num\_items} specifies the number of items.
\end{enumerate}

\subsubsection{Optimization and Regularization}
Our estimation procedure maximizes the log-likelihood in Equation \eqref{eq:sample-loglikelihood} by updating coefficients iteratively using algorithms such as stochastic gradient descent.

\begin{align}
    \text{log-likelihood} = \sum_{n=1}^N \log \hat{P}(i\upn \mid u\upn, s\upn) =  \sum_{n=1}^N \log \frac{\exp(\mu_{u\upn i\upn s\upn})}{\sum_{\ell \in A(s\upn)} \exp(\mu_{u\upn \ell s\upn})}
    \label{eq:sample-loglikelihood}
\end{align}
The \code{torch-choice} package currently supports optimization algorithms in PyTorch, which covers first-order methods such as Adam and quasi-Newton methods like LBFGS.
Since \code{torch-choice} is designed for large-scale datasets and models, second-order methods are not currently supported due to computational constraints.

Researchers often want to regularize magnitudes of coefficients, especially when the number of parameters is large. The \code{torch-choice} implements \code{ConditionalLogitModel} with the support for the commonly used $L^1$ and $L^2$ regularization \citep{krogh1991simple}.
With $L^p$ regularization and regularization weight $\lambda > 0$, the objective function to be maximized changes from log-likelihood to
\begin{align}
    \text{regularized objective function} = \mathop{\mathrm{arg\,max}}_{\bm{\theta}}\underbrace{\sum_{n=1}^N \log \hat{P}(i\upn \mid u\upn, s\upn)}_{\text{log-likelihood}} - \lambda \|\bm{\theta}\|_p
    \label{eq:regularized-loglikelihood}
\end{align}
where $\bm{\theta}$ denotes the set of all coefficients of the model.\footnote{When regularization is used, the model is trained by minimizing the Cross-Entropy loss objective (along with an added regularization term), instead of maximizing an MLE (Maximum Likelihood Estimation) objective, which is what is done without regularization. This is achieved in practice in the package by maximizing the negative of the regularized Cross Entropy objective. The Cross-Entropy objective, without regularization, minimizes the negative likelihood for the choice models we are working with.}

Researchers can add regularization terms by simply specifying \code{regularization} $\in$ \{\code{"L1"}, \code{"L2"}\} and \code{regularization\_weight} $\in \mathbb{R}_{+}$ in the model creation call. As usual, researchers may select the value of the regularization parameter using cross-validation.
The following model initialization code adds $L^1$ regularization with $\lambda = 0.5$ to the model estimation.
\begin{lstlisting}[language=python]
model = ConditionalLogitModel(...,
    regularization="L1", regularization_weight=0.5)
\end{lstlisting}

\subsubsection{Model Estimation}
The \code{torch-choice} package offers a model training pipeline. The following code snippet shows how to train a CLM with the \code{run()} function, in which \code{batch\_size=-1} means the gradient descent is performed on the entire dataset.\footnote{If you don't know what optimizing with mini-batches means, you leave the \code{batch\_size} argument to be -1, which is the default setting.}  We use the LBFGS optimizer in this example since we are working on a small dataset with only 2,779 choice records and 13 coefficients to be estimated. We recommend using the Adam optimizer (the default optimizer) instead for larger datasets and more complicated models.

\begin{lstlisting}[language=python]
from torch_choice import run
run(model, dataset, batch_size=-1, learning_rate=0.003, num_epochs=50_000, model_optimizer="Adam")
\end{lstlisting}
The \code{run(...)} function reports the model provided, the estimation progress, the final log-likelihood, and table summarizing the estimated coefficients. Due to the limited space, we only present the coefficient estimation table in this paper; readers can refer to the supplementary Jupyter Notebook for the complete output.


\begin{lstlisting}[
    basicstyle=\ttfamily\fontsize{7pt}{7pt}\selectfont,
    breaklines=false,
    numbers=none,
    frame=single,
    columns=fullflexible,
    literate={_}{{\_}}1   % <-- auto-escape “_” in this listing only
]
Time taken for training: 313.3691020011902
Skip testing, no test dataset is provided.
==================== model results ====================
Log-likelihood: [Training] -1874.638427734375, [Validation] N/A, [Test] N/A

| Coefficient                           |   Estimation |   Std. Err. |       z-value |    Pr(>|z|) | Significance   |
|:--------------------------------------|-------------:|------------:|--------------:|------------:|:---------------|
| itemsession_cost_freq_ovt[constant]_0 | -0.0372967   |  0.00709548 |  -5.25641     | 1.46895e-07 | ***            |
| itemsession_cost_freq_ovt[constant]_1 |  0.0934432   |  0.00509636 |  18.3353      | 0           | ***            |
| itemsession_cost_freq_ovt[constant]_2 | -0.0427786   |  0.00322258 | -13.2747      | 0           | ***            |
| session_income[item]_0                | -0.0862388   |  0.0183147  |  -4.70874     | 2.49259e-06 | ***            |
| session_income[item]_1                | -0.0269096   |  0.00384866 |  -6.99195     | 2.71094e-12 | ***            |
| session_income[item]_2                | -0.0370623   |  0.00406378 |  -9.12016     | 0           | ***            |
| itemsession_ivt[item-full]_0          |  0.0593741   |  0.010087   |   5.88622     | 3.95136e-09 | ***            |
| itemsession_ivt[item-full]_1          | -0.00635264  |  0.00428462 |  -1.48266     | 0.138165    |                |
| itemsession_ivt[item-full]_2          | -0.00583005  |  0.00189437 |  -3.07757     | 0.00208699  | **             |
| itemsession_ivt[item-full]_3          | -0.00138013  |  0.00118705 |  -1.16265     | 0.24497     |                |
| intercept[item]_0                     |  0.000137248 |  1.26895    |   0.000108159 | 0.999914    |                |
| intercept[item]_1                     |  1.32624     |  0.703773   |   1.88447     | 0.0595013   |                |
| intercept[item]_2                     |  2.81955     |  0.618261   |   4.56045     | 5.10447e-06 | ***            |
Significance codes: 0 '***' 0.001 '**' 0.01 '*' 0.05 '.' 0.1 ' ' 1
\end{lstlisting}

\begin{figure}[h]
    \centering
    \includegraphics[width=0.8\linewidth]{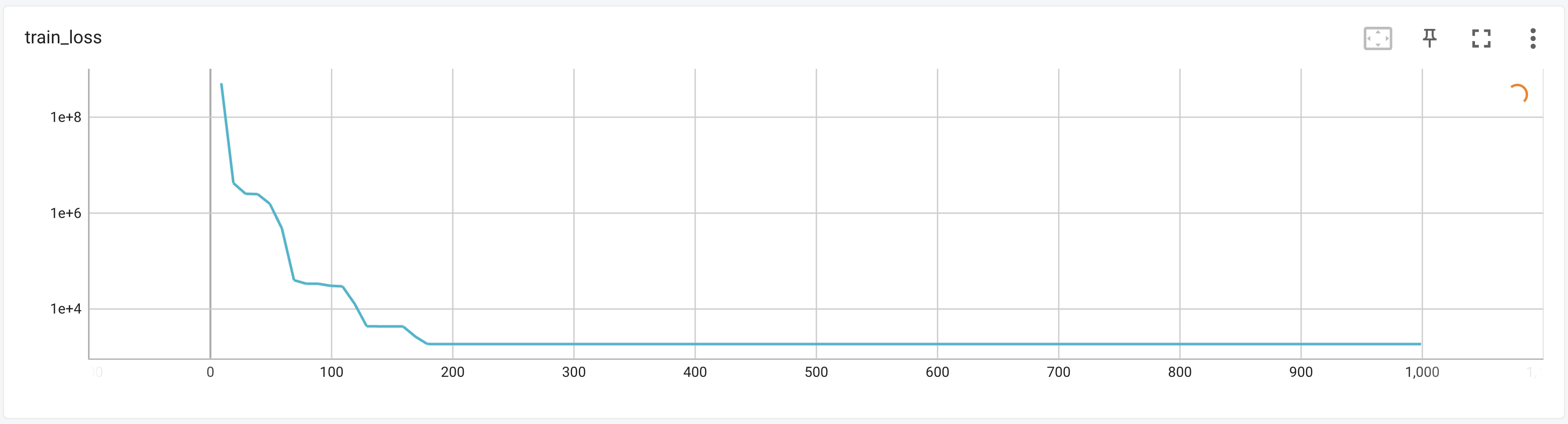}
    \caption{Training Progress Curve in TensorBoard}
    \label{fig:tensorboard-example}
\end{figure}

After the model estimation begins, \code{torch-choice} creates a TensorBoard log to illustrate the training progress; researchers can start a TensorBoard using the following command and navigate to \code{http://localhost:6006} in the browser to obtain training progress; Figure \ref{fig:tensorboard-example} shows an example of the training progress curve, the x-axis is the number of epochs, and the y-axis is the log-likelihood.
Please note that Figure \ref{fig:tensorboard-example} is only an illustrative example, the actual training progress curve may vary depending on the dataset and model configuration.

\begin{lstlisting}[language=bash]
tensorboard --logdir ./lightning_logs --port 6006
\end{lstlisting}

\subsubsection{Post-Estimation}
The \code{torch-choice} package provides utilities to help researchers easily retrieve estimated coefficients from fitted models.
Consider the following stylized model configured with the code block.

\begin{equation}
    \mu_{uis} = \alpha + \beta_i + \gamma_u + \delta_i^\top \textbf{x}^{(user)}_u + \eta^\top \textbf{y}^{(item)}_i + \theta_u^\top \textbf{z}^{(session)}_{s} + \kappa_i^\top \textbf{w}^{(itemsession)}_{is} + \iota_u^\top \textbf{w}^{(itemsession)}_{is}
    \label{eq:clm-post-estimation-example}
\end{equation}

\begin{lstlisting}[language=Python]
model = ConditionalLogitModel(
    formula="(1|constant) + (1|item) + (1|user) + (user_obs|item) + (item_obs|constant) + (session_obs|user) + (itemsession_obs|item) + (itemsession_obs|user)",
    dataset=dataset,
    num_users=num_users,
    num_items=num_items)
\end{lstlisting}

After the model estimation, the \code{get_coefficient()} method allows researchers to retrieve the coefficient values from the model using the general syntax \code{model.get_coefficient(COEFFICIENT_NAME)} with \code{COEFFICIENT_NAME} from the \code{formula} string.
For example,
\begin{itemize}
    \item \code{model.get_coefficient("intercept[constant]")} will return the estimated value of $\alpha$, which is a scalar.
    \item \code{model.get_coefficient("intercept[user]")} returns the array of $\gamma_u$'s, which is a 1D array of length \code{num_users}.
    \item \code{model.get_coefficient("session_obs[user]")} returns the corresponding coefficient $\theta_u$, which is a 2D array of shape \code{(num_users, num_session_features)}. Each row of the returned tensor corresponds to the coefficient vector of a user.
\end{itemize}

Researchers can refer to our online documentation \href{https://gsbdbi.github.io/torch-choice/conditional_logit_model_mode_canada/}{Tutorial: Conditional Logit Model on ModeCanada Dataset} for a complete Jupyter Notebook tutorial.

%% file: sections/nested_logit_model.tex
\subsection{Nested Logit Model}
When items can naturally be clustered into nests, the nested logit model becomes a suitable option.\footnote{A set of nests is a partition of item sets. When the model requires the user to choose only one item across all partitions each time, these partitions are called nests. When such a requirement is removed, the model allows users to choose an item from every partition, these partitions are referred to as categories.}
The nested logit model assumes a weaker version of independence of irrelevant alternatives (IIA): the IIA assumption holds for each pair of items belonging to the same nest but not for pairs from different nests.

\subsubsection{Model Specification}
We implement a two level nested logit model, where at the higher nesting level the user chooses one of many nests, and at the lower level, the user chooses an item within a nest. Thus, the user selects exactly one item across all nests. In the current version of the software, the nested logit is available only in a setting where there is a single category.\footnote{Our nested logit setup can be constrasted with modeling multiple categories using the multinomial logit in \code{torch-choice}, where we model choice over multiple categories simultaneously, and the user chooses one item from each category simultaneously. In the future we aim to support modeling choice over multiple categories where each category can include a two level nested logit.}  The nested logit model decomposes the utility of choosing item $i$ into the item-specific values and inclusive values for each nest as in Equation~(\ref{eq:nlm-utility-decomposition}). This effectively allows correlation among the $\varepsilon_{uis}$ within the choices in a nest.
Nests are indexed by $k \in \{1, 2, \dots, K\}$, $\mathcal{K}(i) \in \{1, 2, \dots, K\}$ denotes the nest of item $i$, and $I_k \subseteq \{1, 2, \dots, I\}$ denotes the set of items in nest $k$.
The \code{NestedLogitModel} constructor requires a \code{nest\_to\_item} dictionary mapping $k \mapsto I_k$, where keys of \code{nest\_to\_item}  are nest indices $k$'s and  \code{nest\_to\_item[j]} is a list consisting of item indices in $I_k$.
Researchers do \emph{not} need to supply a \code{num\_items} argument to the \code{NestedLogitModel} class, the number of items is automatically inferred from the \code{nest\_to\_item} dictionary.

This section briefly discusses the derivation of the log-likelihood formula \code{torch-choice} maximizes. Researchers can refer to Discrete Choice Methods with Simulation by Kenneth Train for a more detailed treatment \citep{train2009discrete}.

Let $\mathcal{V}_{uis} = \mu_{uis} + \varepsilon_{uis}$ denote the utility user $u$ gains from item $i$ in the context of session $s$. The NLM allows $\varepsilon_{uis}$ of items from the same nest to be correlated, but $\varepsilon_{uis}$ of items from different categories should be independent.
The $\lambda_k$ denotes the degree of independence among $\varepsilon_{uis}$'s of items in nest $k$. 
Equation \eqref{eq:nlm-likelihood-original} illustrates the likelihood of item $i$ from the nested logit model.
\begin{align}
    P(i\mid u, s) = \frac{e^{\mu_{uis} / \lambda_{\mathcal{K}(i)}} \left(\sum_{j \in I_{\mathcal{K}(i)}} e^{\mu_{ujs} / \lambda_{\mathcal{K}(i)}}\right)^{\lambda_{\mathcal{K}(i)}-1}}{\sum_{k=1}^K \left(\sum_{j \in I_k} e^{\mu_{ujs} / \lambda_k}\right)^{\lambda_{k}}}
    \label{eq:nlm-likelihood-original}
\end{align}

Without loss of generality, the utility $\mathcal{V}_{uis}$ can be decomposed into a nest-level component $W_{u \mathcal{K}(i) s}$ and an item-level component $T_{uis}$.

\begin{equation}
    \mathcal{V}_{uis} = \mu_{uis} + \varepsilon_{uis} = W_{u \mathcal{K}(i) s} + T_{uis} + \varepsilon_{uis}
    \label{eq:nlm-utility-decomposition}
\end{equation}

The likelihood in Equation \eqref{eq:nlm-likelihood-original} decomposes into a marginal probability of choosing nest $\mathcal{K}(i)$ and a conditional probability of choosing $i$ given nest $\mathcal{K}(i)$ is chosen.
\begin{align}
    P(i\mid u, s) = \underbrace{\frac{\exp\left(T_{uis} / \lambda_{\mathcal{K}(i)}\right)}{\sum_{j \in I_{\mathcal{K}(i)}} \exp\left(T_{ujs} / \lambda_{\mathcal{K}(i)} \right)}
    }_\text{conditional prob. of item}
    \underbrace{\frac{\exp\left(W_{u\mathcal{K}(i)s} + \lambda_{\mathcal{K}(i)} \mathcal{I}_{u \mathcal{K}(i) s} \right)}{\sum_{k=1}^K \exp\left(W_{uks} + \lambda_k \mathcal{I}_{uks}\right)}}_\text{marginal prob. of nest}
    \label{eq:nlm-likelihood-decomposition}
\end{align}
The \emph{inclusive value} of nest $k$ (for user $u$ in session $s$), denoted as $\mathcal{I}_{uks}$, is defined as $\log \sum_{j \in I_k} \exp(T_{ujs}/\lambda_k)$, which is the expected utility from choosing the best alternative from nest $k$ given Gumbel error term in utility.

The softmax form of marginal and conditional probabilities in Equation \eqref{eq:nlm-likelihood-decomposition} suggests a way to model $\mathcal{V}$: estimate both $W_{u \mathcal{K}(i) s}$ and $T_{uis}$ using linear functional forms as in the conditional logit model with nest observables and item observables, respectively.

\begin{align}
    W_{uks} &= \sum_{p=1}^P \bm{\theta}_p^\top \bm{x}_p \label{eq:nest-W} \\
    T_{uis} &= \sum_{p=1}^{P'} \bm{\gamma}_p^\top \bm{z}_p \label{eq:nest-T}
\end{align}
where $\bm{\theta}_p$'s are coefficients of the nest-level model that can be (1) fixed for all items and users, (2) user-specific, and (3) item-specific (i.e., nest-specific since we are talking about the nest level model). $\bm{x}_p$'s are nest level observables.
The nest-level model in NLM is similar to a CLM for choices over categories as items.
The only difference between specifying the nest-level model and a CLM is that an \emph{item-specific} coefficient or observable in the NLM nest-level model is indeed \emph{nest-specific}. For example, by specifying \code{nest\_formula="(1|item)"} for the nest-level model, the model specification in Equation \eqref{eq:nest-W} is $W_{uks} = \lambda_c$.
$\bm{\gamma}_p$'s denote coefficients of item-level model and $\bm{z}_p$'s are item observables. Specification of the item-level model is exactly the same as in the CLM.

The configuration of NLM requires specifying both the item-level CLM and nest-level CLM.
Similar to the CLM, researchers can configure models using Python dictionaries or R-like formulas.
For example,
\begin{lstlisting}[language=python]
nest_formula = '(1|item) + (item_obs1|constant) + (session_obs1|user) + (user_obs1|item)'
item_formula = '(1|user) + (item_obs2|user) + (session_obs2|item) + (user_obs2|constant)'
\end{lstlisting}
specify the following model
\begin{align}
    W_{uks} &= \tau_k + \beta^\top \bm{x}^{(\text{item obs 1})}_k + \gamma^\top_u \bm{x}^{(\text{session obs 1})}_s + \delta^\top_k \bm{x}^{(\text{user obs 1})}_u \label{eq:W-example}\\
    T_{uis} &= \alpha_u + \tau_u^\top \bm{x}^{(\text{item obs 2})}_i + \phi^\top_i \bm{x}^{(\text{session obs 2})}_s + \psi^\top \bm{x}^{(\text{user obs 2})}_u \label{eq:Y-exaple}
\end{align}
It is worth mentioning that everything tagged with \code{item} in the \code{nest\_formula}, such as item-specific intercepts/coefficients (i.e., \code{(1|item)} and \code{(user\_obs1|item)}) and item-specific observables (i.e., \code{(item\_obs1|constant)}) are indeed nest-specific with $k$ sub-script in Equation \eqref{eq:W-example}. For example, by specifying nest-level formula \code{"(1|item)+(user\_obs1|item)"}, it is interpreted as \emph{nest-specific fixed effect} together with \emph{nest-specific random effect from user observable 1}. Equivalently, $W_{uks} = \tau_k + \delta^\top_k \bm{x}^{(\text{user obs 1})}_u$.

Similar to specifying CLMs, researchers can use \code{\{nest, item\}\_coef\_variation\_dict} and \code{\{nest, item\}\_num\_param\_dict} to specify the functional form of $W_{uks}$ and $T_{uis}$ respectively.

Due to the decomposition in Equation \eqref{eq:nlm-likelihood-decomposition}, the log-likelihood for user $u$ to choose item $i$ in session $s$ can be written as Equation \eqref{eq:nested-likelihood}, and \code{torch-choice} jointly optimizes all coefficients to maximize Equation \eqref{eq:nested-likelihood}.

\begin{equation}
    \begin{aligned}
    \log P(i \mid u, s) &= \log P_\text{item}(i \mid u, s) + \log P_\text{nest}(\mathcal{K}(i) \mid u, s) \\
    &= \log \left(\frac{\exp(T_{uis}/\lambda_k)}{\sum_{j \in I_{\mathcal{K}(i)}} \exp(T_{ujs}/\lambda_k)}\right) + \log \left( \frac{\exp(W_{u\mathcal{K}(i)s} + \lambda_{\mathcal{K}(i)} \mathcal{I}_{u\mathcal{K}(i)s})}{\sum_{k=1}^K \exp(W_{uks} + \lambda_k \mathcal{I}_{uks})}\right)
    \end{aligned}
\label{eq:nested-likelihood}
\end{equation}

\code{torch-choice} allows for empty nest-level models from an empty dictionary or empty formula string. In this case, $W_{uks} = \varepsilon_{uks}$ since the inclusive value term $\lambda_k I_{uks}$ will be used to model the choice over categories.
In this special case, the model's log-likelihood function reduces to Equation~(\ref{eq:nested-likelihood-no-nest}).
\begin{equation}
    \log P(i \mid u, t) = \log \left(\frac{\exp(T_{uit}/\lambda_k)}{\sum_{j \in B_k} \exp(T_{ujt}/\lambda_k)}\right) + \log \left( \frac{\lambda_k \mathcal{I}_{ukt}}{\sum_{\ell=1}^K \lambda_\ell \mathcal{I}_{u\ell t}}\right)
    \label{eq:nested-likelihood-no-nest}
\end{equation}
However, by specifying an empty item-level model ($T_{uis} = \varepsilon_{uis}$), the nested logit model reduces to a conditional logit model of choices over categories. Hence, researchers should never use the \code{NestedLogitModel} class with an empty item-level model.

The $\lambda_k$'s capture the degree of independence among unobserved utilities $\varepsilon$ of items in nest $k$ \citep{train1978goods}.
\code{torch-choice} estimates $\lambda_k$'s jointly with other coefficients. By default, these paramaters differ over categories, indicating different levels of correlations among unobserved utilities of items in different categories.
Researchers can specify \code{shared\_lambda=True} to constrain $\lambda_k = \lambda$ for all categories, indicating the same level of correlations in all categories.
Whether to add such a constraint is an empirical question depending on specific datasets; researchers can use hypothesis testing to determine the right configuration \citep{train2009discrete}.
While working with large datasets, setting the same $\lambda$ for all categories helps reduce the number of parameters when the number of categories is potentially very large.

The following code snippets demonstrate the construction of nested logit models in \code{torch-choice}.
\begin{lstlisting}[language=python]
# Creation of NestedLogitModel using formulas.
model = NestedLogitModel(
    nest_to_item=nest_to_item,
    nest_formula='(1|item) + (item_obs1|constant) + (session_obs1|user) + (user_obs1|item)',
    item_formula='(1|user) + (item_obs2|user) + (session_obs2|item) + (user_obs2|constant)',
    dataset=dataset,
    shared_lambda=True)
\end{lstlisting}
Similar to the CLMs, researchers can add $L^1$ or $L^2$ regularizations to NLMs by specifying \code{regularization} $\in$ \{\code{"L1"}, \code{"L2"}\}and \code{regularization\_weight} $\in \mathbb{R}_{+}$ keywords.
For example, the following code creates a model that penalizes $\|\bm{\theta}\|_2$ during estimation.
\begin{lstlisting}[language=python]
model = NestedLogitModel(regularization="L2", regularization_weight=1.5)
\end{lstlisting}

\subsubsection{Dataset Preparation}
The \code{NestedLogitModel} admits a \code{JointDataset} structure, which consists of separate \code{ChoiceDataset} objects holding observables for categories and items.
Naming conventions for \code{ChoiceDataset} for item observables are exactly the same as before: researchers can specify item observables \code{item\_<obs\_name>}, user observables \code{user\_<obs\_name>}, session observables \code{session\_<obs\_name>}, and (item, session)-specific observables \code{item\_session\_<obs\_name>}.
For the dataset holding nest information, researchers need to add the \code{item\_index} array $(i\upn)_{n=1}^N$, and all nest-specific observables should be named as \code{item\_<obs\_name>} because categories are considered as items in the nest-level model. Similarly, researchers can add (nest, session)-specific observables via the \code{item\_session\_<obs\_name>} keyword.
Lastly, the researcher can link two \code{ChoiceDataset} objects using the \code{JointDataset} class as in the following code block.

\begin{lstlisting}[language=python]
# create the nest-level dataset.
nest_dataset = ChoiceDataset(item_index=item_index, item_obs=..., use_obs=...)
# create the item-level dataset.
item_dataset = ChoiceDataset(item_index=item_index, item_obs=..., use_obs=...)
# create the joint dataset.
dataset = JointDataset(nest=nest_dataset, item=item_dataset)
\end{lstlisting}

\subsubsection{Optimization and Model Estimation}
For NLM estimation, \code{torch-choice} maximizes the likelihood defined in Equation \eqref{eq:nlm-likelihood-original} by updating all coefficients iteratively using optimization algorithms like stochastic gradient descent.
To estimate NLM, researchers need to provide a joint dataset encompassing both nest and item-level datasets. The \code{run()} helper function works for NLM and joint datasets.

\begin{lstlisting}[language=python]
from torch_choice import run
# create the nest-level dataset.
nest_dataset = ChoiceDataset(item_index=item_index.clone())
# create the item-level dataset.
item_dataset = ChoiceDataset(item_index=item_index, price_obs=price_obs)
# create the joint dataset.
dataset = JointDataset(nest=nest_dataset, item=item_dataset)
# estimate the model.
run(model, dataset, learning_rate=0.01, num_epochs=5000, model_optimizer="Adam")
\end{lstlisting}
Similar to the conditional logit model, researchers can specify regularization on coefficients by passing \code{regularization} and \code{regularization_weight} into the \code{run()} method.

Researchers can refer to our online documentation here - \href{https://gsbdbi.github.io/torch-choice/nested_logit_model_house_cooling/}{Random Utility Model (RUM) Part II: Nested Logit Model}
 for complete tutorials on applying various NLM to the \emph{Heating and Cooling System Choice in Newly Built Houses in California} dataset \citep{train2009discrete}.

\subsubsection{Post-Estimation}
The nested logit model has a similar interface for coefficient extraction to the conditional logit model demonstrated above.
Consider a nested logit model consisting of an item-level model in Equation \eqref{eq:clm-post-estimation-example} and a nest-level model incorporating user-fixed effect, category-fixed effect (i.e., specified by \code{(1|item)} in the \code{nest_formula}), and the user-specific coefficient on a 64-dimensional nest-specific observable (i.e., specified by \code{(item_obs|user)} in the \code{nest_formula}).

\begin{lstlisting}[language=python]
nested_model = NestedLogitModel(
    nest_to_item=nest_to_item,
    nest_formula='(1|user) + (1|item) + (item_obs|user)',
    item_formula='(1|constant) + (1|item) + (1|user) + (user_obs|item) + (item_obs|constant) + (session_obs|user) + (itemsession_obs|item) + (itemsession_obs|user)',
    num_users=num_users,
    dataset=joint_dataset,
    shared_lambda=False)
\end{lstlisting}

Researchers need to retrieve the coefficients of the nested logit model using the \code{get_coefficient()} method with an additional \code{level} argument.

For example, researchers can use the following code snippet to retrieve the coefficient of the user-fixed effect in the nest-level model, which is a vector with \code{num_users} elements.
\begin{lstlisting}[language=python]
nested_model.get_coefficient("intercept[user]", level="nest")
\end{lstlisting}
With \code{level="item"}, the researcher can obtain the coefficient of user-specific fixed effect in the item level model (i.e., $\gamma_u$'s in Equation \eqref{eq:clm-post-estimation-example}), which is a vector with length \code{num_users}.
\begin{lstlisting}[language=python]
nested_model.get_coefficient("intercept[user]", level="item")
\end{lstlisting}
Such API generalizes to all other coefficients listed above, such as \code{itemsession_obs[item]} and \code{itemsession_obs[user]}.
One exception is the coefficients for inclusive values (often denoted as $\lambda$). Researchers can retrieve the coefficient of the inclusive value by using \code{get_coefficient("lambda")} without specifying the \code{level} argument.
In fact, \code{get_coefficient} will disregard any \code{level} argument if the coefficient name is \code{lambda}.
The returned value is a scalar if \code{shared_lambda=True}, and a 1D array of length \code{num_nests} if \code{shared_lambda=False}.

%% file: sections/benchmark.tex
\section{Performance Benchmarks}\label{section:performance}
Leveraging parallel processors in modern graphical processing units (GPU), the \code{torch-choice} package is particularly efficient dealing with a large amount of parameters. For example, estimating the item-specific slope $\beta_i^{(2)}$ in Equation \eqref{eq:clm-example-formula} becomes computationally expensive when we have a large number of items. This section examines the computational efficiency of \code{torch-choice} as the dataset grows larger in different dimensions.\footnote{Through this section, \code{torch-choice} uses full-batch Adam optimization with a learning rate of 0.01 to maximize the sample likelihood of the dataset. The optimization stops at an epoch when the sample likelihood fails to improve for more than 50 epochs. The optimization runs for up to 5,000 epochs to ensure full algorithm convergence. Experiments related to \code{torch-choice} are run on an Nvidia RTX3090 GPU. R benchmarks were conducted on a server with 16 cores and 128GiB of memory.}

For the analysis in this section, we create a synthetic seed dataset of consumers' choices; then, we subsample datasets of various sizes to measure the runtime performance of packages.

\subsection{Synthetic Data Generation}
\paragraph{User Features}

Let $\mathbf{x}_u \in \mathbb{R}^{d}$ denote the $d$-dimensional feature vector for user $u \in \{1,2,\ldots,U\}$, where we simulate $U = 500$ users and $d = 30$ features. User features $\bm{x}_u$ are generated from a mixture of $K_u = 5$ Gaussian clusters:
\begin{align}
    \textbf{x}_u \sim \sum_{k=1}^{K_u} \pi_k^u \mathcal{N}(\mathbf{c}_k^u, \sigma_u^2 \mathbf{I})
\end{align}
where $\sigma_u = 0.5$ is the within-cluster standard deviation.
Cluster centers are sampled from a Gaussian distribution $\textbf{c}_k^u \sim \mathcal{N}(\mathbf{0}, \delta_u^2\mathbf{I})$ with between cluster distance $\delta_u = 3$.
Cluster assignments $\pi_k^u$ are determined by uniform allocation with random perturbation so that the $\{\pi_1^u, \dots, \pi_{K_u}^u\}$ vector is one-hot for each user $u$.
We selected $K_u = 5$ to represent distinct market segments with clear separation ($\delta_u = 3$) while maintaining heterogeneity within cluster ($\sigma_u = 0.5$). This parameterization yields well-formed clusters.

\paragraph{Item Features}
Item features are generated analogously to these user features.
Let $\mathbf{z}_i \in \mathbb{R}^{d}$ denote the feature vector for item $i \in \{1,2,\ldots,I\}$, we simulated $I = 500$ items with $d = 30$ item-specific features.
\begin{align}
    \mathbf{z}_i \sim \sum_{k=1}^{K_i} \pi_k^i \mathcal{N}(\mathbf{d}_k^i, \sigma_i^2 \mathbf{I})
\end{align}
where $\sigma_i = 0.3$ is the within-cluster standard deviation. Cluster centers $\mu_k^i \sim N(\mathbf{0}, \delta_i^2 \mathbf{I})$ with $\delta_i = 2.5$. The cluster assignments $\pi_k^i$ are determined by uniform allocation with random perturbation and each item belongs to a single cluster.
The smaller within-cluster variance ($\sigma_i = 0.3$) and between-cluster variance ($\delta_i = 2.5$) reflect that product categories are typically more tightly defined than consumer segments.
The 30-dimensional feature space was chosen to capture the complex preference structure while remaining computationally feasible.

\paragraph{User Price Sensitivity Coefficients} In our simulation, each user $u$ has a price sensitivity coefficient $\beta_u$ generated as:
\begin{align}
    \beta_u \sim N(\mu_\beta, \sigma_\beta^2)
\end{align}
where $\mu_\beta = -1.5$ and $\sigma_\beta = 0.5$.

\paragraph{Item Prices}
For each item $i$ in session $s \in \{1,2,\ldots,S\}$, where $S = 1,000$, prices $p_{s,i}$ are generated from the following steps.
We first generate a base-price $p_i^\text{base} = \max(\eta_i, 1.0)$ with  $\eta_i \sim \mathcal{N}(10, 3^2)$ for each item $i$. This base price is constant across sessions.
In each session $s$, the price of item $i$, $p_{s, i}$, is constructed by adding symmetric noise with much smaller variance $\varepsilon_{s,i} \sim \mathcal{N}(0, (0.6)^2)$ to the base-price: $p_{i, s} = p_i^\text{base} + \varepsilon_{s,i}$.
Finally, we take $p_{s, i} = \max(p_{s, i}, 1)$ to make sure all prices are positive.

\paragraph{Item Availabilities}
We assume a 95\% availability of items in our simulation. Specifically, item availability $A_{s,i} \in \{0,1\}$ in session $s$ is:

\begin{align}
    A_{s,i} \sim \text{Bernoulli}(0.95)
\end{align}

with the constraint that $\sum_i A_{s,i} \geq 1$ for all $s$ (i.e., the item is at least available in one session).

\paragraph{Choice Generation Process}
We generate $3,000,000$ choice records, each associating a randomly selected (i.e., sampling from uniform distributions) user $u^{(n)}$, session $s^{(n)}$. The corresponding chosen item $i^{(n)}$ is generated using a hybrid choice model combining exploration and exploitation to mimic real-world consumer choices.
The utility of item $i$ for user $u^{(n)}$ in session $s^{(n)}$ is:

\begin{align}
    \mathcal{V}_{u^{(n)},i,s^{(n)}} = \textbf{x}_{u^{(n)}}^\top \textbf{z}_i + \beta_{u^{(n)}} \cdot p_{s^{(n)}, i} + \varepsilon_{u^{(n)},i,s^{(n)}}
\end{align}
where $\varepsilon_{u_n,i,s_n} \sim \mathcal{N}(0, 0.5^2)$ is a random utility shock.

With probability 0.15, the user makes a purely exploratory choice by randomly picking an available item:
\begin{align}
    i^{(n)} \sim \text{Uniform}(\{i : A_{s^{(n)},i} = 1\})
\end{align}

Otherwise, with a probability of 0.85, we further create two nested cases.
With probability 0.35 ($0.85 \times 0.35 = 0.2975$ total probability), the item chosen is sampled from a softmax with temperature $T = 2.0$:

\begin{align}
    P(i^{(n)} = i) = \frac{\exp(\mathcal{V}_{u_n,i,s^{(n)}/T)}}{\sum_{j:A_{s^{(n)},j}=1} \exp(\mathcal{V}_{u^{(n)},j,s^{(n)}}/T)}
\end{align}
With probability 0.65 ($0.85 \times 0.35 = 0.5525$ total probability), choice follows the maximum utility rule:

\begin{align}
    i^{(n)} = \text{argmax}_{i:A_{s^{(n)},i}=1} \mathcal{V}_{u^{(n)},i, s^{(n)}}
\end{align}

The exploration rate of 15\% and mixed decision rules capture the dual-process nature of consumer decision-making, from utility maximization to more stochastic choice patterns, ensuring the dataset exhibits realistic choice behavior across various market conditions.

The seed dataset includes 3,000,000 choice records from 500 users and 500 items; each user ($u$) and item ($i$) is associated with a 30-dimensional vector of characteristics generated randomly, denoted as $X_u \in \R^{30}$ and $Z_i \in \R^{30}$ respectively. There is no session-specific observable.
We consider the models in Equations \eqref{eq:benchmark-model-1} to \eqref{eq:benchmark-model-3} to cover model specifications with and without random effects.
\begin{align}
    \mu_{uis} &= \bm{\beta}^\top \textbf{z}_i
    \label{eq:benchmark-model-1} \\
    \mu_{uis} &= \bm{\alpha}_i^\top \textbf{x}_u
    \label{eq:benchmark-model-2} \\
    \mu_{uis} &= \bm{\alpha}_i^\top \textbf{x}_u + \bm{\beta}^\top \textbf{z}_i
    \label{eq:benchmark-model-3}
\end{align}
As a convention, the coefficient of the first item $\bm{\alpha}_0$ is set to zero.

Experiments in this section demonstrate the scalability of \code{torch-choice} when researchers expand (1) the number of choice records (i.e., size of dataset), (2) the dimension of user/item covariates, and (3) the cardinality of the item set.
For each of these, and for each specification above, two sets of experiments are conducted: (1) we compare the performance of \code{torch-choice} and an open source implementation, \code{mlogit} implemented in R, on datasets with small-to-medium sizes;\footnote{The \code{torch-choice} and R comparison only focuses on small-scale datasets because the size of long-format CSV file grows fast as the dataset expands. For example, 200,000 records of 500 items result in 100 million rows of data with 60 columns of user/item characteristics, which takes more than 100GiB of disk space and memory, which leads to out-of-memory issues.}
(2) we then examine the scalability of \code{torch-choice} when we expand to even larger datasets.
We run each of these benchmarks five times and report the average and standard deviation of run times across five trials.

\subsection{Scalability with Number of Records}
The first set of experiments focuses on the performance of \code{torch-choice} as the number of records ($N$) grows while the dimensions of user/item observables and the item set remain the same.

\begin{figure}
    \begin{subfigure}[b]{\linewidth}
        \centering
        \includegraphics[width=\linewidth]{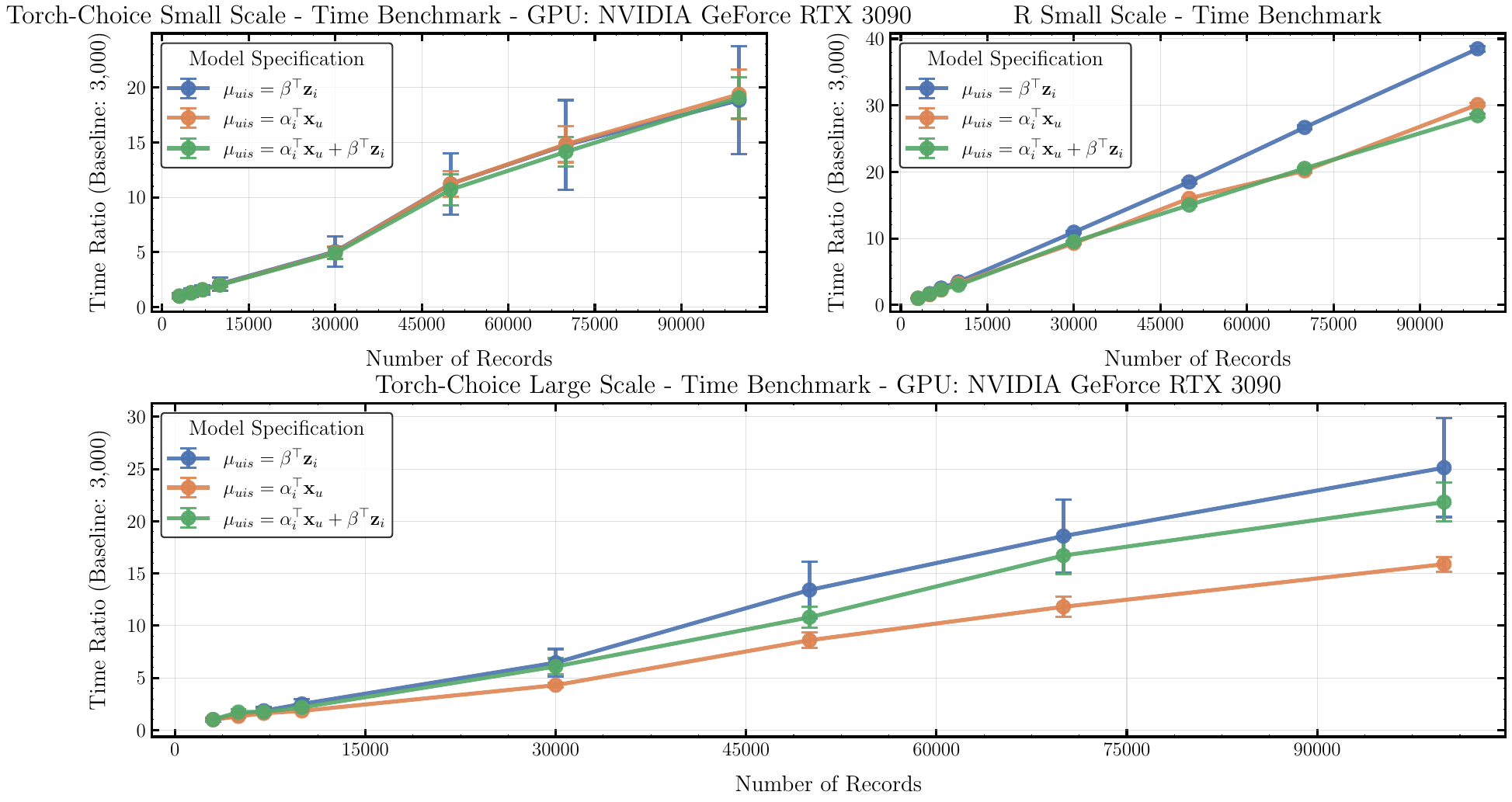}
        \caption{Time Cost Relative to Baseline Case}
        \label{fig:num-records-comparison}
    \end{subfigure}
    \begin{subfigure}[b]{\linewidth}
        \centering
        \includegraphics[width=\linewidth]{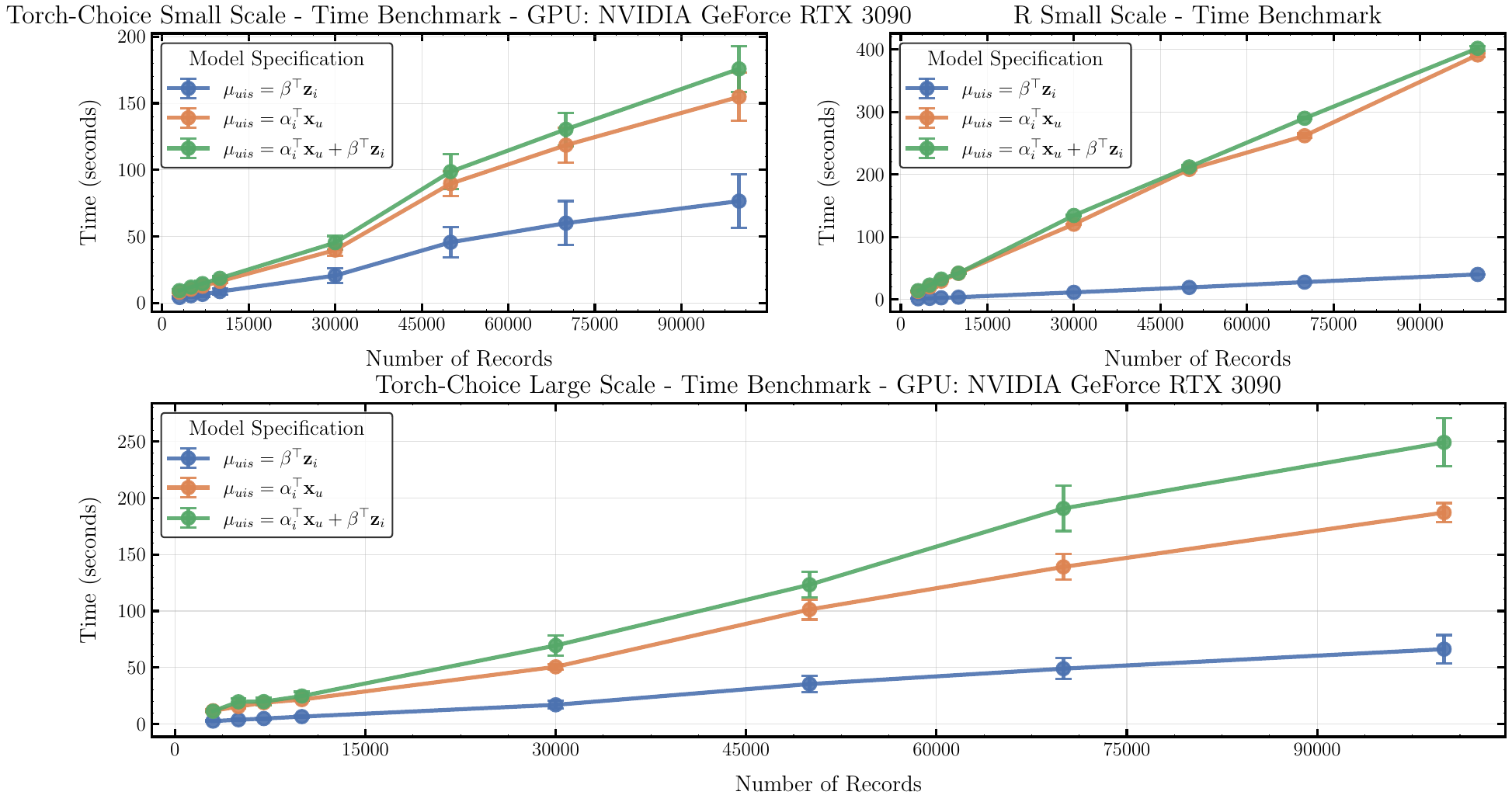}
        \caption{Time Cost in Seconds}
        \label{fig:abs-time-num-records-comparison}
    \end{subfigure}
    \caption{Comparison of Computational Time Costs as the Number of Records $N$ Increases}
\end{figure}

\textbf{Scalability on Small Datasets.}
We sampled $30$ items and only used the first 10 dimensions in user and item characteristics through this experiment.
The top panels in Figure \ref{fig:num-records-comparison} show the time cost of \code{torch-choice}  as the number of records $N$ grows from 3,000 to 100,000. The vertical axis shows the ratio between the time taken at each level of $N$ and the baseline time cost at $N=3,000$.
Both \code{torch-choice} and \code{mlogit} scale linearly as $N$ grows, but the runtime of the R-implementation increases by 30x to 40x as the sample size rises from 3,000 to 100,000 while \code{torch-choice} only increases by 20x.

\textbf{Scalability on Larger Datasets.}
We then turn to the complete dataset with $500$ items and $30$-dimensional user/item characteristics.
The bottom panel in Figure \ref{fig:num-records-comparison} illustrates the time taken relative to the baseline case ($N=3,000$) as $N$ increases from $3,000$ to $100,000$, which leads to a 30x increase in run time.
The runtime of model $\mu_{uis} = \bm{\beta}^\top \mathbf{z}_i$ becomes vastly different from the other two because models with random effects now have 500x more parameters than model $\mu_{uis} = \bm{\beta}^\top \mathbf{z}_i$.
The linear cost of \code{torch-choice} is preserved on the large-scale dataset, which provides evidence of scalability.

\textbf{Absolute Time Costs.}
Figure \ref{fig:abs-time-num-records-comparison} depicts the time cost in seconds of \code{mlogit} and \code{torch-choice} on different datasets. The Figure illustrates that in these datasets, \code{torch-choice} provided up to a 2x reduction in run time.

\newpage
\subsection{Scalability with Number of Covariates/Parameters}
The number of parameters plays an important role in computational efficiency; as the number of parameters grows, the time cost of computing the gradient of the likelihood function also grows.  In this section, we examine the performance of \code{torch-choice} as the number of covariates grows.
We investigate the models in Equations \eqref{eq:benchmark-model-1} to \eqref{eq:benchmark-model-3} using only the first $P \in \{3, 5, 10, 15, 20, 30\}$ dimensions of user features $\mathbf{x}_u$ and item features $\mathbf{z}_i$.

\begin{figure}[h!]
    \begin{subfigure}[b]{\linewidth}
        \centering
        \includegraphics[width=\linewidth]{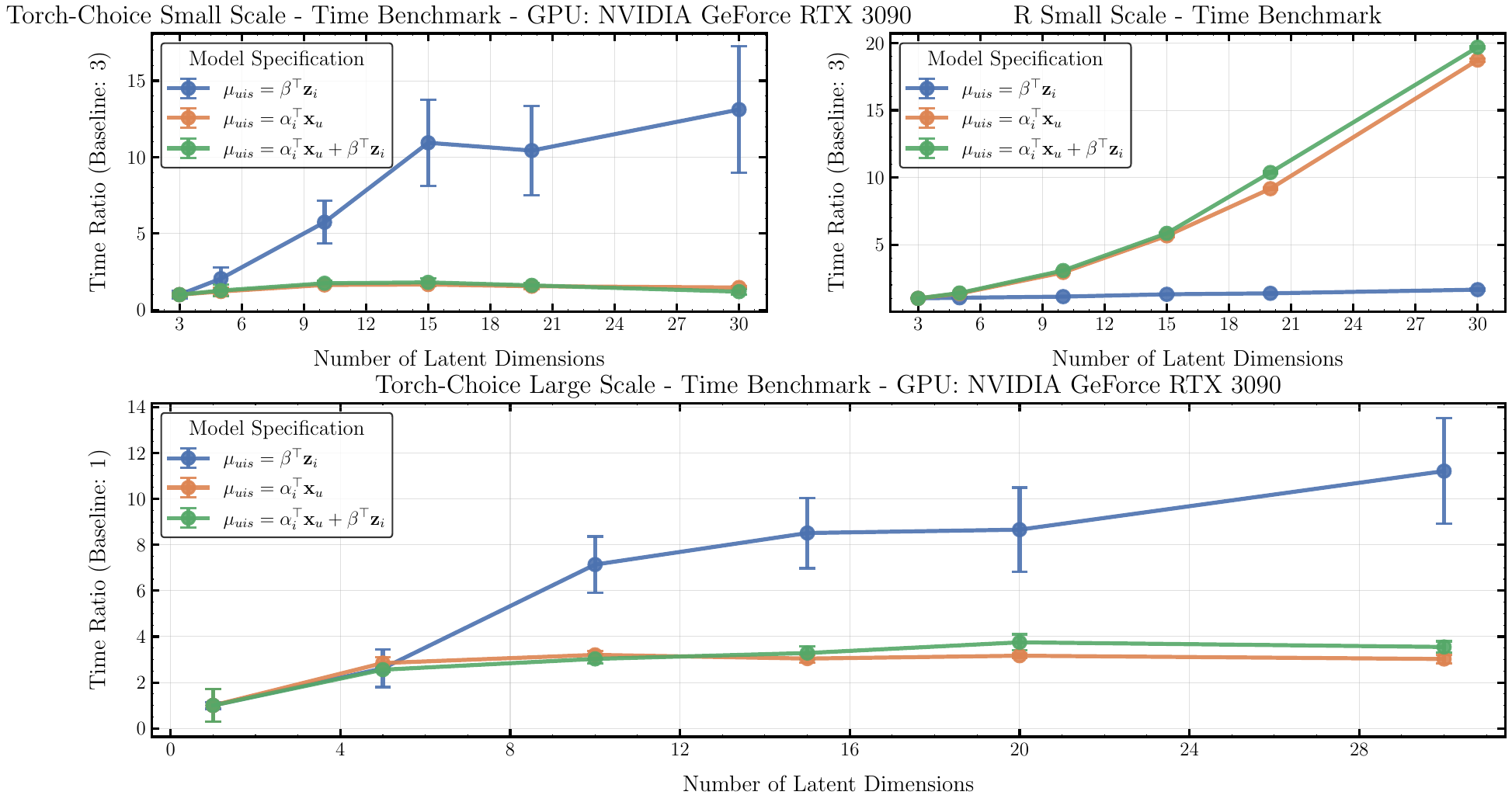}
        \caption{Time Cost Relative to Baseline Case}
        \label{fig:num-covariates-comparison}
    \end{subfigure}
    \begin{subfigure}[b]{\linewidth}
        \centering
        \includegraphics[width=\linewidth]{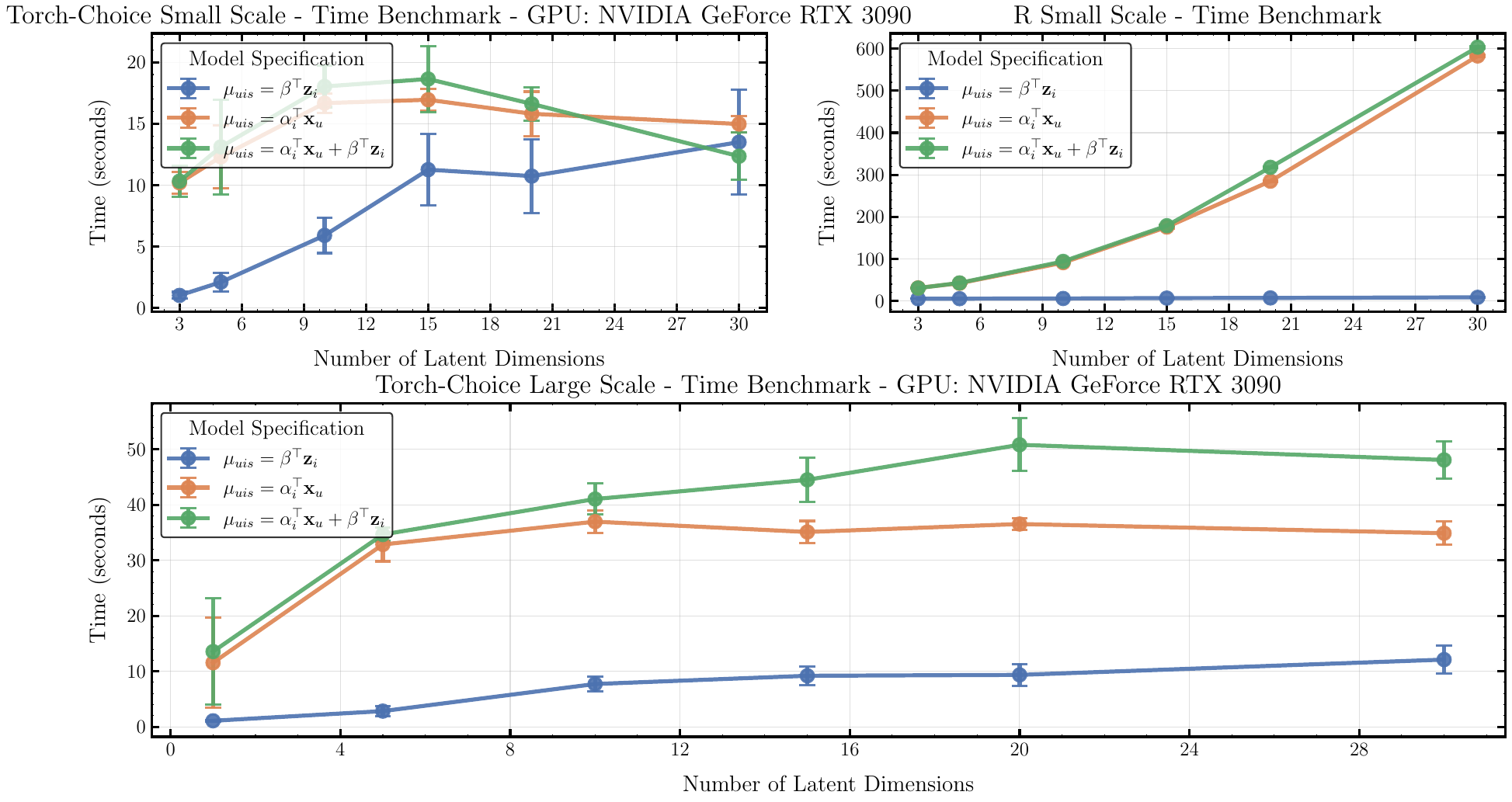}
        \caption{Time Cost in Seconds}
        \label{fig:abs-time-num-covariates-comparison}
    \end{subfigure}
    \caption{Comparison of Computational Time Costs as the Number of Covariates $P$ Increases}
\end{figure}

\textbf{Scalability on Small Datasets.}
The first part examines different packages' time costs at each level of $P$ normalized by the baseline runtime ($P=3$) on a dataset with $N=10,000$ records and $I=50$ items.
The top panels of Figure \ref{fig:num-covariates-comparison} compare computational efficiencies of R and \code{torch-choice} as the number of covariates grows.
For the two models with item-specific coefficients $\alpha_i$, the time taken is more than 20x higher when $P=30$ compared to the baseline case while using the R-implementation; in contrast, the time cost was less than 3x higher when $P=30$ with \code{torch-choice}.
The time ratio for the $\beta^\top \mathbf{z}_i$ model using \code{torch-choice} is higher because the model converges extremely fast in the baseline setting, which can be verified in Figure \ref{fig:abs-time-num-covariates-comparison}.

\textbf{Scalability on Larger Datasets.}
The bottom panel of Figure \ref{fig:num-covariates-comparison} illustrates the run time with different $P$'s on the large-scale dataset with 3 million records, $500$ users/items. Model $ \mu_{uis} = \bm{\alpha}_i^\top X_u + \bm{\beta}^\top Z_i$ with both random and common coefficients scaled worse than linear scaling, the run time increased by up to 4x when $P=30$.
The scaling factor in time cost as $P$ expands depends on the number of items, which we will explore later, partly because the total number of parameters is $\mathcal{O}(P\times I)$. To summarize, \code{torch-choice} scales well when the number of items is moderate. Researchers should use caution while specifying item-specific random effects on high dimensional covariates when many items are present.

\textbf{Absolute Time Costs.}
Finally, Figure \ref{fig:abs-time-num-covariates-comparison} compares time cost in seconds of R and \code{torch-choice} in various scenarios, in which \code{torch-choice} delivers up to 30x speed-up compared to \code{mlogit}.

\newpage
\subsection{Scalability with Number of Items}
The number of parameters grows as the item set expands as well when item-specific coefficients are specified. In this section, we examine the performance of \code{torch-choice} as the number of items grows.

\begin{figure}[h!]
    \centering
    \begin{subfigure}[b]{\linewidth}
        \centering
        \includegraphics[width=\linewidth]{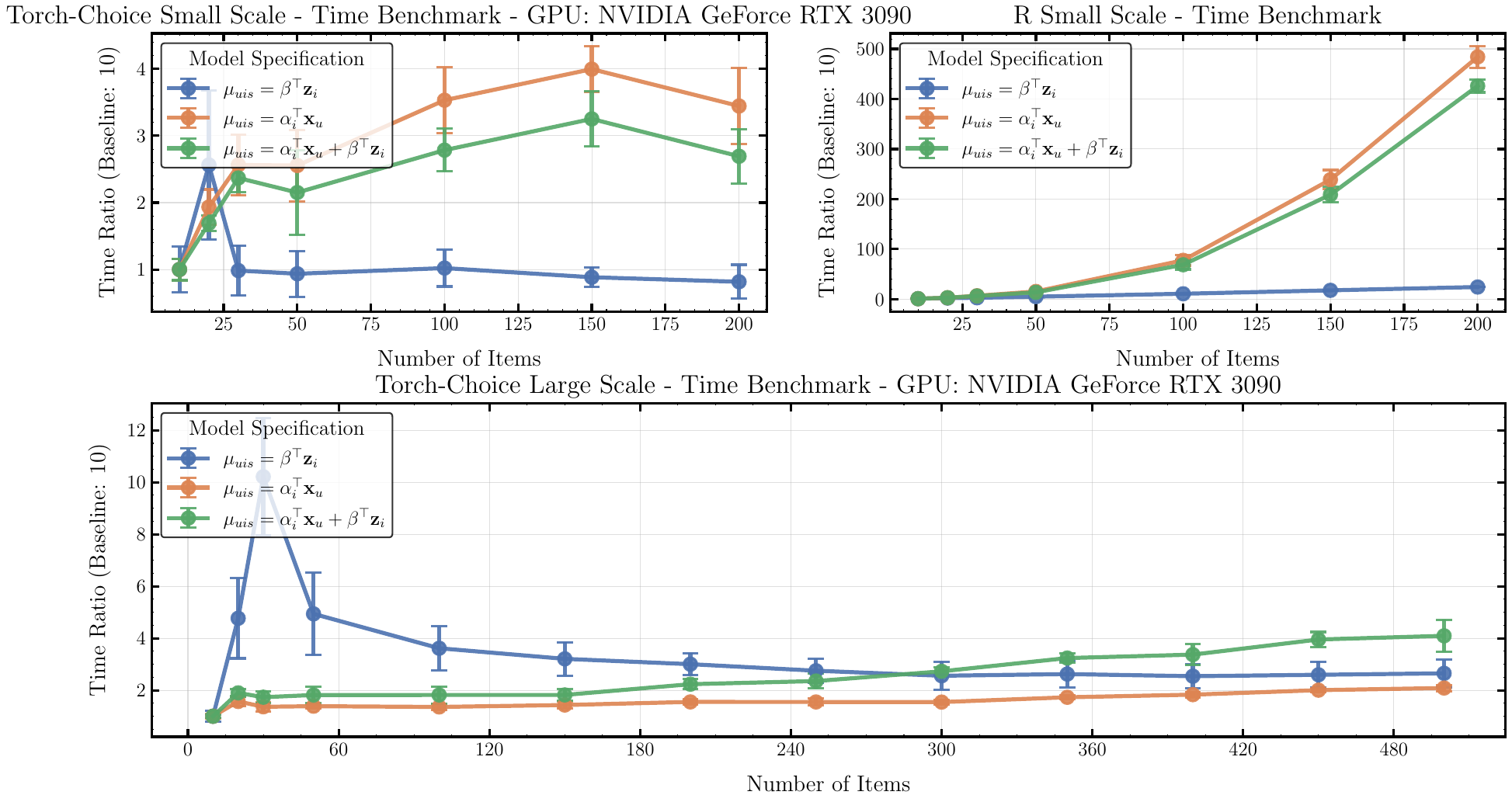}
        \caption{Time Cost Relative to Baseline Case}
        \label{fig:num-items-comparison}
    \end{subfigure}
    \vspace{0.5cm}
    \begin{subfigure}[b]{\linewidth}
        \centering
        \includegraphics[width=\linewidth]{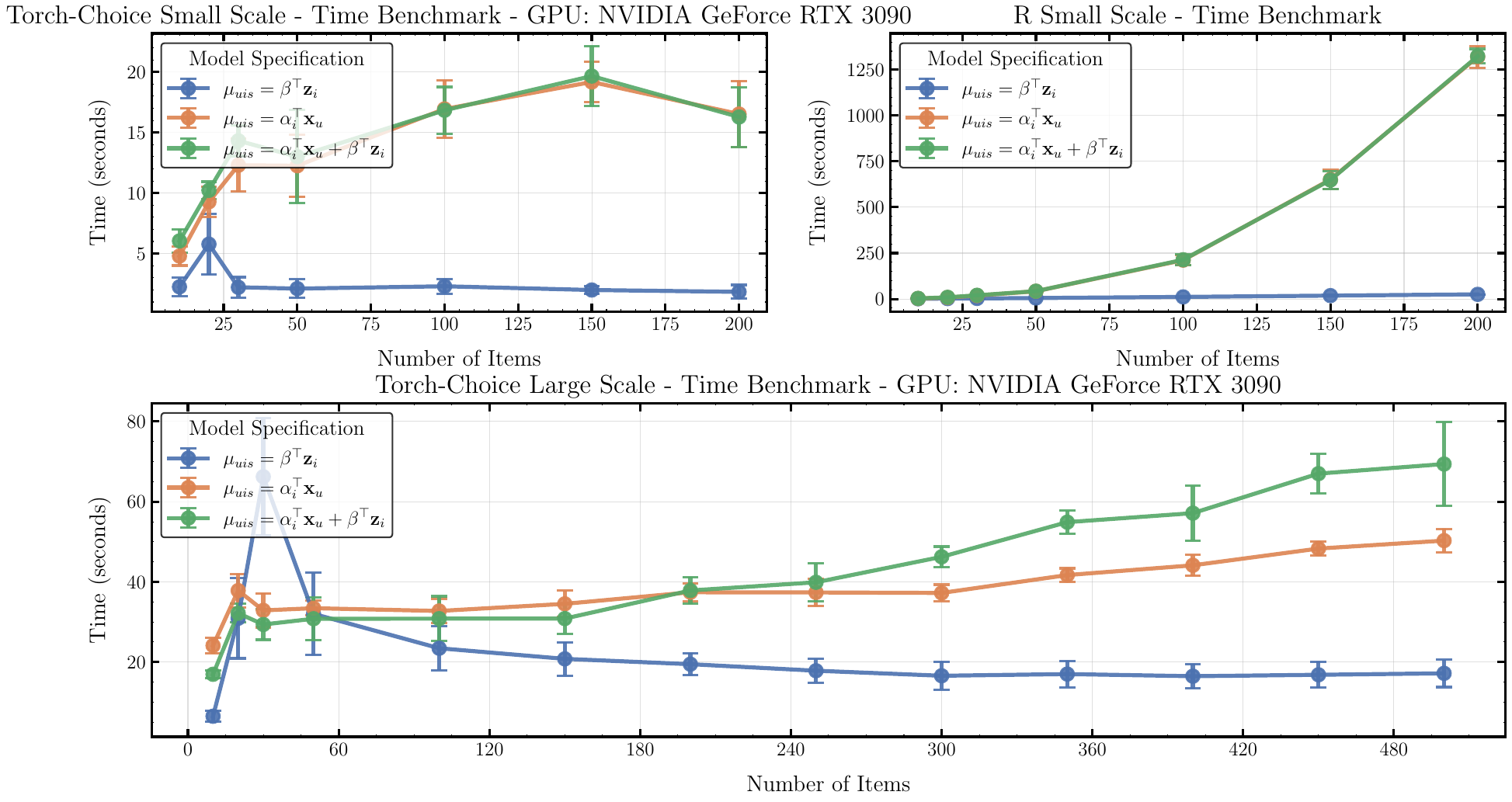}
        \caption{Time Cost in Seconds}
        \label{fig:abs-time-num-items-comparison}
    \end{subfigure}
    \caption{Comparison of Computational Time Costs as the Number of Items $I$ Increases}
    \label{fig:num-item-benchmark}
\end{figure}

\textbf{Scalability on Small Datasets.}
We first compare performances, in terms of time cost relative to baseline case ($I=10$), of \code{torch-choice} and R-implementation as $I$ increases from $10$ to $200$ on a dataset with $N=10,000$ records and 5-dimensional user/item characteristics.\footnote{We choose a smaller $N$ in this experiment because we have a limited number of observations when $I=10$ and we wish to keep $N$ constant when $I$ changes.}\footnote{We stop at $I=200$ because \code{mlogit} encountered an out-of-memory error with $I=250$ on a server with 128GiB memory.}
We investigate the models in Equations \eqref{eq:benchmark-model-1}, \eqref{eq:benchmark-model-2}, and \eqref{eq:benchmark-model-3}.
The top panels in Figure \ref{fig:num-items-comparison} show the time cost of \code{torch-choice} and \code{mlogit} as the number of items $I$ grows.
Since the complexity of model $\mu_{uis} = \bm{\beta}^\top \mathbf{z}_i$ does not depend on the number of items, its runtime remains relatively constant as the number of items grows.
For the other two models, \code{torch-choice} exhibits a mild runtime increase as $I$ grows, and the costs of estimating these models are only 3x-4x higher when the number of items increases by 20x from 10 to 200.
In contrast, the curvatures in the panel for R implementation suggest an exponential growth trend in the time cost of \code{mlogit}; the computational cost required explodes by more than 400x-500x when the number of items rises by 20x.
These observations again highlight the computational efficiency of parallel computing capability in PyTorch and \code{torch-choice} while handling a large number of parameters.

\textbf{Scalability on Larger Datasets.}
The bottom panel in Figure \ref{fig:num-items-comparison} demonstrates the runtime of \code{torch-choice} as $I$ increases up to 500 with $N=30,000$ and all 30 user/item features.
As we expected, the cost of estimating model $\mu_{uis} = \bm{\beta}^\top \mathbf{z}_i$ grows slowly as we expand the item set.
The computational costs of both model $\mu_{uis} = \bm{\alpha}_i^\top \mathbf{x}_u$ and model $\mu_{uis} = \bm{\alpha}_i^\top \mathbf{X}_u + \bm{\beta}^\top \mathbf{z}_i$ rise around 4x as we expand the item set. This provides further evidence that the computational cost using \code{torch-choice} follows a sub-linear trend with the expansion of the item set, demonstrating \code{torch-choice}'s scalability.

\textbf{Absolute Time Costs.}
Lastly, Figure \ref{fig:abs-time-num-items-comparison} demonstrates the run time of \code{torch-choice} and \code{mlogit} in seconds through experiments considered in this section, showing superior performance of \code{torch-choice}. Especially, we observe up to 60x speed-up from the \code{torch-choice} compared to \code{mlogit} for models with item-specific coefficients.

%% file: sections/conclusion.tex
\section{Conclusion and Future Work}
Choice modeling problems are ubiquitous in multiple disciplines, including Economics, Psychology, Education and Computer Science. As a result, access to effective choice modeling software is crucial for both researchers and practitioners. The \code{torch-choice} package aims to greatly improve this modeling landscape by providing a flexible and scalable choice modeling implementation.

In this paper, we proposed \code{torch-choice}, a PyTorch-based choice modeling package. 
The \code{torch-choice} package offers a memory-efficient data structure, \code{ChoiceDataset} for storing choice datasets. Towards this end, \code{torch-choice} also provides functionalities allowing researchers to build the \code{ChoiceDataset}, from databases in various formats. For the advanced user, the \code{ChoiceDataset} object is completely flexible and supports all functionalities in PyTorch's dataset data structure.

The \code{torch-choice} package implements two widely-used classes of models, the \code{ConditionalLogitModel} and the \code{NestedLogitModel} using \code{PyTorch}.
It allows researchers to fully specify flexible functional forms and changing item availability, which overcomes some of the difficulties when modeling with existing libraries.
Models in \code{torch-choice} support regularization during model estimation, making it usable for engineering applications as well, a feature not provided in standard choice modeling libraries used by econometricians and statisticians. Such a feature vastly increases the scope of choice modeling options as compared to standard libraries used by engineers in terms of flexibility and scale.

The package has the advantage of scaling to massive datasets while being computationally efficient by using GPUs for estimation.
We examined the scalability of \code{mlogit} in R and \code{torch-choice} while (1) increasing the number of observations, (2) increasing the number of covariates, and (3) expanding the item set with various model specifications. Experiment results show that \code{torch-choice} scales significantly better than R on small-to-medium-scale datasets. Then we test \code{torch-choice} on large-scale datasets that R cannot handle, on which \code{torch-choice} demonstrates its ability to scale up with computational efficiency.

Moving forward, we aim to maintain the reference implementation of \code{torch-choice} on the Python Package Index (PyPI) via Github, developing relatively few new features.
We foresee the following important extensions to our models. First, \code{torch-choice} currently does not allow researchers to specify Bayesian priors over coefficients and learn posteriors instead of point estimates. With Bayesian coefficients, in addition to using item/user characteristics as observables, researchers can incorporate these characteristics into the coefficient prior and let them influence the coefficient posterior directly.
Second, currently our nested logit model only supports two levels of nesting.
Another direction is to implement multi-level nested logit models (i.e., nests within a nest), which are particularly useful when users make sequential decisions. In addition, allowing for modeling multiple categories with nested logit will be an important extension.
We are actively developing these extensions and aim to release them soon.

%% file: sections/acknowledgements.tex
\section*{Acknowledgements}

We want to express our sincere appreciation and gratitude to the following researchers who have contributed to the completion of this paper.
We thank Charles P\'ebereau, Keshav Agrawal, Aaron Kaye, Rob Kuan, Emil Palikot, Rachel Zhou, Rob Donnelly, and Undral Byambadalai for their valuable comments and feedback that help improve the quality of our package and paper. 

%% file: ms.bbl
\begin{thebibliography}{10}

\bibitem{JSSv095i11}
Yves Croissant.
\newblock Estimation of random utility models in r: The mlogit package.
\newblock {\em Journal of Statistical Software}, 95(11):1–41, 2020.

\bibitem{berry1995}
Steven Berry, James Levinsohn, and Ariel Pakes.
\newblock Automobile prices in market equilibrium.
\newblock {\em Econometrica: Journal of the Econometric Society}, pages 841--890, 1995.

\bibitem{berry2004}
Steven Berry, James Levinsohn, and Ariel Pakes.
\newblock Differentiated products demand systems from a combination of micro and macro data: The new car market.
\newblock {\em Journal of political Economy}, 112(1):68--105, 2004.

\bibitem{chintagunta2001}
Pradeep Chintagunta, Ekaterini Kyriazidou, and Josef Perktold.
\newblock Panel data analysis of household brand choices.
\newblock {\em Journal of Econometrics}, 103(1-2):111--153, 2001.

\bibitem{mcfadden1974measurement}
Daniel McFadden.
\newblock The measurement of urban travel demand.
\newblock {\em Journal of public economics}, 3(4):303--328, 1974.

\bibitem{wilson1990}
Frank~R Wilson, Sundar Damodaran, and J~David Innes.
\newblock Disaggregate mode choice models for intercity passenger travel in canada.
\newblock {\em Canadian Journal of Civil Engineering}, 17(2):184--191, 1990.

\bibitem{miyamoto2004}
Kazuaki Miyamoto, Varameth Vichiensan, Naoki Shimomura, and Antonio P{\'a}ez.
\newblock Discrete choice model with structuralized spatial effects for location analysis.
\newblock {\em Transportation Research Record}, 1898(1):183--190, 2004.

\bibitem{kornstad2007}
Tom Kornstad and Thor~O Thoresen.
\newblock A discrete choice model for labor supply and childcare.
\newblock {\em Journal of Population Economics}, 20(4):781--803, 2007.

\bibitem{van1995}
Arthur Van~Soest.
\newblock Structural models of family labor supply: a discrete choice approach.
\newblock {\em Journal of human Resources}, pages 63--88, 1995.

\bibitem{baker2001basics}
Frank~B Baker.
\newblock {\em The basics of item response theory}.
\newblock ERIC, 2001.

\bibitem{embretson2013}
Susan~E Embretson and Steven~P Reise.
\newblock {\em Item response theory}.
\newblock Psychology Press, 2013.

\bibitem{bonoma1979}
Thomas~V Bonoma and Wesley~J Johnston.
\newblock Decision making under uncertainty: a direct measurement approach.
\newblock {\em Journal of Consumer Research}, 6(2):177--191, 1979.

\bibitem{fleiss1986}
Joseph~L Fleiss, Janet~BW Williams, and Alan~F Dubro.
\newblock The logistic regression analysis of psychiatric data.
\newblock {\em Journal of Psychiatric Research}, 20(3):195--209, 1986.

\bibitem{zhang2002}
Tong Zhang and Vijay~S Iyengar.
\newblock Recommender systems using linear classifiers.
\newblock {\em The Journal of Machine Learning Research}, 2:313--334, 2002.

\bibitem{chang2008}
Ming-wei Chang, Wen-tau Yih, and Christopher Meek.
\newblock Partitioned logistic regression for spam filtering.
\newblock In {\em Proceedings of the 14th ACM SIGKDD international conference on Knowledge discovery and data mining}, pages 97--105, 2008.

\bibitem{deng2012mnist}
Li~Deng.
\newblock The mnist database of handwritten digit images for machine learning research [best of the web].
\newblock {\em IEEE signal processing magazine}, 29(6):141--142, 2012.

\bibitem{athey2018}
Susan Athey, David Blei, Robert Donnelly, Francisco Ruiz, and Tobias Schmidt.
\newblock Estimating heterogeneous consumer preferences for restaurants and travel time using mobile location data.
\newblock In {\em AEA Papers and Proceedings}, volume 108, pages 64--67, 2018.

\bibitem{donnelly2021}
Robert Donnelly, Francisco~JR Ruiz, David Blei, and Susan Athey.
\newblock Counterfactual inference for consumer choice across many product categories.
\newblock {\em Quantitative Marketing and Economics}, 19(3):369--407, 2021.

\bibitem{ruiz2020}
Francisco~JR Ruiz, Susan Athey, and David~M Blei.
\newblock Shopper: A probabilistic model of consumer choice with substitutes and complements.
\newblock {\em The Annals of Applied Statistics}, 14(1):1--27, 2020.

\bibitem{donnelly2022}
Robert Donnelly, Ayush Kanodia, and Ilya Morozov.
\newblock Welfare effects of personalized rankings.
\newblock {\em Available at SSRN 3649342}, 2022.

\bibitem{vafa2022}
Keyon Vafa, Emil Palikot, Tianyu Du, Ayush Kanodia, Susan Athey, and David~M Blei.
\newblock Learning transferrable representations of career trajectories for economic prediction.
\newblock {\em arXiv preprint arXiv:2202.08370}, 2022.

\bibitem{krogh1991simple}
Anders Krogh and John Hertz.
\newblock A simple weight decay can improve generalization.
\newblock {\em Advances in neural information processing systems}, 4, 1991.

\bibitem{scikit-learn}
F.~Pedregosa, G.~Varoquaux, A.~Gramfort, V.~Michel, B.~Thirion, O.~Grisel, M.~Blondel, P.~Prettenhofer, R.~Weiss, V.~Dubourg, J.~Vanderplas, A.~Passos, D.~Cournapeau, M.~Brucher, M.~Perrot, and E.~Duchesnay.
\newblock Scikit-learn: Machine learning in {P}ython.
\newblock {\em Journal of Machine Learning Research}, 12:2825--2830, 2011.

\bibitem{MicroPyBLP}
Christopher Conlon and Jeff Gortmaker.
\newblock Incorporating micro data into differentiated products demand estimation with {PyBLP}, 2023.
\newblock Working paper.

\bibitem{arteaga2022xlogit}
Cristian Arteaga, JeeWoong Park, Prithvi~Bhat Beeramoole, and Alexander Paz.
\newblock xlogit: An open-source python package for gpu-accelerated estimation of mixed logit models.
\newblock {\em Journal of Choice Modeling}, 42:100339, 2022.

\bibitem{kingma2014adam}
Diederik~P Kingma and Jimmy Ba.
\newblock Adam: A method for stochastic optimization.
\newblock {\em arXiv preprint arXiv:1412.6980}, 2014.

\bibitem{liu1989lbfgs}
Dong~C Liu and Jorge Nocedal.
\newblock On the limited memory bfgs method for large scale optimization.
\newblock {\em Mathematical programming}, 45(1-3):503--528, 1989.

\bibitem{falcon2019pytorch}
William Falcon et~al.
\newblock Pytorch lightning.
\newblock {\em GitHub. Note: https://github. com/PyTorchLightning/pytorch-lightning}, 3(6), 2019.

\bibitem{rudolph2017structured}
Maja Rudolph, Francisco Ruiz, Susan Athey, and David Blei.
\newblock Structured embedding models for grouped data.
\newblock {\em Advances in neural information processing systems}, 30, 2017.

\bibitem{Kluyver2016jupyter}
Thomas Kluyver, Benjamin Ragan-Kelley, Fernando P{\'e}rez, Brian Granger, Matthias Bussonnier, Jonathan Frederic, Kyle Kelley, Jessica Hamrick, Jason Grout, Sylvain Corlay, Paul Ivanov, Dami{\'a}n Avila, Safia Abdalla, and Carol Willing.
\newblock Jupyter notebooks -- a publishing format for reproducible computational workflows.
\newblock In F.~Loizides and B.~Schmidt, editors, {\em Positioning and Power in Academic Publishing: Players, Agents and Agendas}, pages 87 -- 90. IOS Press, 2016.

\bibitem{mcfadden1973conditional}
Daniel McFadden et~al.
\newblock Conditional logit analysis of qualitative choice behavior.
\newblock 1973.

\bibitem{bhat1995heteroscedastic-modecanada}
Chandra~R Bhat.
\newblock A heteroscedastic extreme value model of intercity travel mode choice.
\newblock {\em Transportation Research Part B: Methodological}, 29(6):471--483, 1995.

\bibitem{train2009discrete}
Kenneth~E Train.
\newblock {\em Discrete choice methods with simulation}.
\newblock Cambridge university press, 2009.

\bibitem{train1978goods}
Kenneth Train and Daniel McFadden.
\newblock The goods/leisure tradeoff and disaggregate work trip mode choice models.
\newblock {\em Transportation research}, 12(5):349--353, 1978.

\end{thebibliography}
